\documentclass{bmvc2k}

\title{Face Aging via Diffusion-based Editing}

\addauthor{Xiangyi Chen}{xiangyi.chen@telecom-paris.fr}{1,2}
\addauthor{Stéphane Lathuilière}{stephane.lathuiliere@telecom-paris.fr}{2}

\addinstitution{
 Shanghai Jiao Tong University \\
 Shanghai, China
}
\addinstitution{
 LTCI, Télécom Paris \\
 Institut Polytechnique de Paris \\
 Palaiseau, France
}

\runninghead{Chen, Lathuilière}{Face Aging via Diffusion-based Editing}

\def\ie{\emph{i.e}\bmvaOneDot}
\def\eg{\emph{e.g}\bmvaOneDot}

\def\etal{\emph{et al}\bmvaOneDot}
\def\method{FADING}

\begin{document}

\maketitle

\begin{abstract}

In this paper, we address the problem of \textit{face aging}—generating past or future facial images by incorporating age-related changes to the given face. Previous aging methods rely solely on human facial image datasets and are thus constrained by their inherent scale and bias. This restricts their application to a limited generatable age range and the inability to handle large age gaps. We propose FADING, a novel approach to address \textbf{F}ace \textbf{A}ging via \textbf{DI}ffusion-based editi\textbf{NG}. We go beyond existing methods by leveraging the rich prior of large-scale language-image diffusion models. First, we specialize a pre-trained diffusion model for the task of face age editing by using an age-aware fine-tuning scheme. Next, we invert the input image to latent noise and obtain optimized null text embeddings. Finally, we perform text-guided local age editing via attention control. The quantitative and qualitative analyses demonstrate that our method outperforms existing approaches with respect to aging accuracy, attribute preservation, and aging quality.
\end{abstract}

\section{Introduction}

Have you ever looked in the mirror and wondered what you might look like in a few decades? Digital face aging techniques make it possible now. This exciting field aims to create realistic transformations of a person's face, simulating the effects of aging or de-aging. These techniques have critical applications in various fields including entertainment, forensics, and healthcare.  
A number of face-aging methods have been introduced. Most recent approaches~\cite{gomez2022custom,yao2021high,alaluf2021only,or2020lifespan,makhmudkhujaev2021re,he2021disentangled,he2019s2gan} are based on deep generative models, such as generative adversarial networks (GANs)~\cite{goodfellow2020generative} and have shown promising results. But to our knowledge, all existing learning-based methods rely solely on datasets of human facial images  (\eg~FFHQ~\cite{karras2019style} or CelebA~\cite{liu2015deep, karras2017progressive}), and are thus constrained by the inherent scale and bias of these datasets. For example, most methods have a limited transformation range (mostly less than 70  years old) and may fail when faced with large age gaps, occlusions, as well as extreme head poses due to the limited number of these rare cases in the dataset. 

Meanwhile, the recently proposed diffusion models~\cite{ho2020denoising} exhibit comparable or even superior generation quality compared to GANs. In light of this, we propose to extend a diffusion-based large-scale language image model to tackle the specific task of face aging.
Our motivation is that these models have learned, through language supervision, a rich image prior on a vast diversity of images, including faces, and have extensive semantic knowledge on diverse concepts (such as \textit{"woman"}/\textit{"man"}, \textit{"glasses"}, etc) that could be potentially exploited for age editing.
While some recent research~\cite{hertz2022prompt,meng2021sdedit,avrahami2022blended,avrahami2022blended2,couairon2022diffedit,kawar2022imagic} has explored the potential of leveraging diffusion models for image editing tasks, they are limited to general-purpose editing methods.  In contrast, no studies have demonstrated how these approaches can be adapted to tailor highly specific tasks such as face aging.

To this end, we propose \method~: Face Aging via DIffusion-based editiNG. The proposed method consists of two stages: specialization and editing. Specialization is a training stage where we re-target a pre-trained diffusion-based language-image model for face aging. In this stage, we employ an age-aware fine-tuning scheme that achieves better disentanglement of the age from age-irrelevant features (\eg gender). For the editing stage, we first employ a well-chosen inversion technique to invert the input image into latent noise. Subsequently, we leverage a pair of text prompts containing both initial and target age information to perform text-based localized age editing, via attention control.
Our contribution can be summarized as follows: 
(i) \method~is the first method to extend large-scale diffusion models for face aging; 
(ii) we successfully leverage the attention mechanism for accurate age manipulation and disentanglement;
(iii) we qualitatively and quantitatively demonstrate the superiority of \method~over state-of-the-art methods through extensive experiments. \footnote{Code available at \url{https://github.com/MunchkinChen/FADING}.}

\section{Related Work}
\paragraph{Face-Aging}
Most of the recent methods rely on the well-known Generative Adversarial Networks (GANs)~\cite{goodfellow2020generative}. 
On the one hand, \textit{condition-based} methods follow the conditional GAN framework~\cite{mirza2014conditional}. This means they include age as an extra condition into the GAN framework to guide age-aware synthesis~\cite{antipov2017face, zhang2017age,li2021continuous, hsu2021wasserstein,wang2018face}. The age estimator can be embedded into the generator and trained simultaneously with it~\cite{li2021continuous}.  
Alternatively, recurrent neural networks are used in \cite{wang2016recurrent,wang2018recurrent} to iteratively synthesize aging effects.
Pre-trained face recognizers are employed to preserve age-irrelevant features (\ie identity)  \cite{antipov2017face,yang2019learning, hsu2021wasserstein,wang2018face}.

On the other hand, other methods~\cite{he2019s2gan, yao2021high, makhmudkhujaev2021re, he2021disentangled, huang2021age}  resort to \textit{latent space manipulation}~\cite{harkonen2020ganspace,shen2020interpreting}. 
An age modulation network is designed to fuse age labels with the latent vectors in HRFAE~\cite{yao2021high}, or to output age-aware transformation to apply to the decoder in RAGAN~\cite{makhmudkhujaev2021re}. SAM~\cite{alaluf2021only} relies on the latent space of a pre-trained GAN and employs an age regressor to explicitly guide the  encoder in generating age-aware latent codes. Huang \etal \cite{huang2021age} learn a unified embedding of age and identity.
Some works also adopt a style-based architecture~\cite{karras2019style, karras2020analyzing}. LATS~\cite{or2020lifespan} follows StyleGAN2~\cite{karras2020analyzing} to perform modulated convolutions to inject learned age code into the decoder.  CUSP~\cite{gomez2022custom} disentangles style and content representations and uses a decoder to combine the two representations with a style-based strategy. We highlight that one drawback of these methods is the significant discrepancy in identity that arises when real images are inverted into the GAN’s latent space ~\cite{tov2021designing}. Consequently, the reconstruction of the initial image may be inaccurate, which can lead to suboptimal results.


\paragraph{Image editing with Diffusion Models (DMs)}
Large-scale diffusion models have raised the bar for text-to-image synthesis \cite{saharia2022photorealistic, ramesh2022hierarchical, rombach2022high}. Naturally, works have attempted to adapt text-guided diffusion models to image editing. SDEdit \cite{meng2021sdedit} is among the first to propose diffusion-based image editing. It adds noise to the input image and then performs a text-guided denoising process from a predefined step. However, SDEdit lacks specific control over edited region. With the help of a mask provided by the user, \cite{avrahami2022blended,avrahami2022blended2,nichol2021glide} better address this problem and enable more meaningful local editing. After each denoising step, the mask is applied to the latent image while also adding the noisy version of the original image. 
DiffEdit \cite{couairon2022diffedit} gets rid of the need for a user-provided mask by automatically generating one that highlights regions to be edited  based on the text description. Prompt-to-prompt \cite{hertz2022prompt} proposes a text-only editing technique based on a pair of \textit{"before-after"} text descriptions. Null-text inversion \cite{mokady2022null} enables real image editing with prompt-to-prompt thanks to its accurate inversion of real images. Concurrently, Imagic \cite{kawar2022imagic} enables text-guided real image editing by fine-tuning the diffusion model to capture the input image's appearance. However, it is important to note that all these methods are general-purpose editing techniques. As such, our work aims to showcase the potential for adapting these broad approaches for use in more specific tasks, such as face aging.

\begin{figure*}
     \centering
     \begin{subfigure}[b]{0.35\textwidth}
         \centering
         \includegraphics[height=3cm]{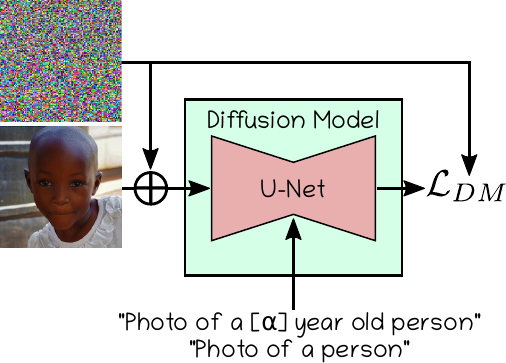}
         \caption{Specialization to aging via finetuning of a pre-trained diffusion model.}
     \end{subfigure}
     \hfill
     \begin{subfigure}[b]{0.62\textwidth}
         \centering
         \includegraphics[height=3.2cm]{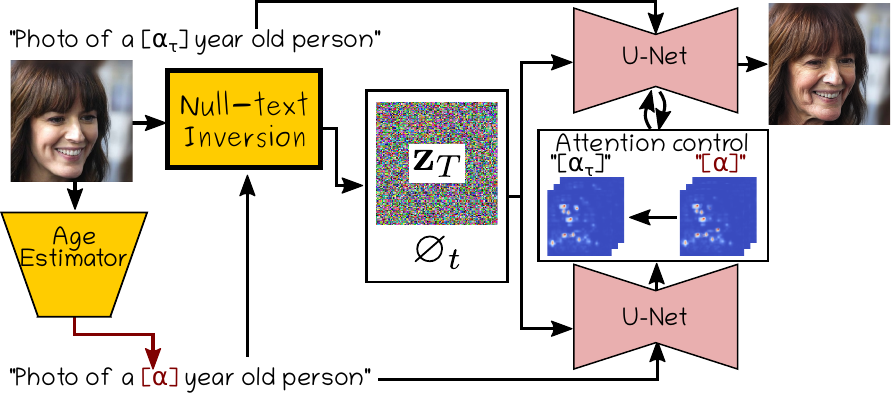}
         \caption{Age editing: given an input image, the diffusion process is inverted. The image is then edited replacing the estimated age with the target age.}

     \end{subfigure}
            \caption{\method~addresses \textbf{f}ace \textbf{a}ging via \textbf{di}ffusion-based editi\textbf{ng}: In the specialization stage, a pre-trained diffusion model is fine-tuned for the aging task. Editing is achieved via age estimation, image inversion, and attention control. 
        }
        \label{fig:pipeline}
        \vspace{-4mm}
\end{figure*}

\section{FADING: Face Aging via DIffusion-based editiNG}


The objective of this work is to transform an input image $\xmat$ to make the person in the image appear to be of a specific target age $\alpha_{\tau}$. For this, we employ a dataset of $N$ face images $\xmat^{(n)}\!\in\!\mathbb{R}^{H\times W\times3},\ n=1,...,N$ with their corresponding age labels $\alpha^{(n)}\in \{1,..K\}$, where $K$ is the maximum age in our training dataset. The age labels $\alpha^{(n)}$ can be obtained either via manual labeling or using a pre-trained age classifier. 

The proposed approach relies on a specialization and an edition stage illustrated in Figure~\ref{fig:pipeline}. In the first stage, a pre-trained diffusion model is re-targeted for the task of face age editing. This training procedure is detailed in Sec.~\ref{sec:adaptation}. To better disentangle age information from other age-irrelevant features, our specialization procedure employs an age-aware fine-tuning scheme. Then, our inference consists of two steps: inversion and editing. 
In the inversion step, we inverse the diffusion process using a recent optimization-based inversion~\cite{mokady2022null} as detailed in Sec.~\ref{sec:inversion}. In the editing step, we use a new prompt that contains the target age to guide a localized age editing with attention control (see Sec.~\ref{sec:editing}). 
We also provide a solution to improve the prompts used for editing to achieve higher image quality. 

\subsection{Specialization to Face Aging}
\label{sec:adaptation}
\method~leverages a pre-trained text-to-image Diffusion Model (DM) \cite{ho2020denoising}.
While the proposed method could be applied to any text-to-image DM, in our experiments, we employ a variant of DM named Latent Diffusion Model (LDMs) \cite{rombach2022high}. LDMs operate in the latent space of an auto-encoder to achieve lower computation complexity. As traditional DMs, LDMs are composed of a forward and a backward pass.

In the forward process, the input image $\xmat_0$ is projected to the auto-encoder latent space, $\zmat_0 = \mathcal{E}(\xmat_0)$. Then, random Gaussian noises are added to the original latent embedding $\zmat_0$ in a stepwise manner to create a sequence of noisy samples ($\zmat_1...,\zmat_{T}$). Learning an LDM consists in  training a neural network $\epsilon_\theta$ to estimate the corresponding noise from a given sample $\zmat_t$. In the reverse process, on the other hand, new data points are generated by sampling from a normal distribution and gradually denoising the sample using $\epsilon_\theta$.
The generated image $\hat\xmat_0$ is obtained by feeding the estimated latent tensor $\hat\zmat_0$ to the decoder.
To enable generation conditioned on a text prompt $\mathcal{P}$, a sequence of token embeddings is extracted from  $\mathcal{P}$ and given to $\epsilon_\theta$ via cross-attention layers, where keys and values are estimated from the token embedding. In the case of unconditional generation, the token embeddings are replaced by fixed embeddings referred to as \textit{null-text embedding} and denoted by $\varnothing_t$. 

Age editing with a pre-trained DM can be performed without any training stage \cite{meng2021sdedit,mokady2022null}, but this produces unsatisfactory results since they are generally not specialized for human faces. Also, coarse conditioning prompts, such as "\textit{man in his thirties}",  can capture age-related semantics but we observe that they often fail to capture more specific textual descriptions of age as numbers, such as "\textit{32-year-old man}". 
To address these issues, we propose a specialization stage that re-purposes a pre-trained DM toward the aging task.
For every face image $\xmat$ with its corresponding age $\alpha$, fine-tuning is performed using an image-prompt pair, with the following prompt: $\mathcal{P}_{\alpha}$="\textit{photo of a $[\alpha]$ year old person}", where $\alpha$ is the age of the person written as numerals. We have observed better performance when adding another age-agnostic prompt $\mathcal{P}$="\textit{photo of a person}" at every iteration. We refer to this fine-tuning scheme as the \textit{double-prompt} scheme.
One assumption to justify this observation is that it can allow better disentangling of age information from other age-irrelevant features (\ie identity and context features). Regarding the training loss, we employ the reconstruction objective of DMs which, in our case, can be written as follows:
\begin{equation}
    \mathcal{L}_{DM}=\mathbb{E}_{\zmat_0\sim\mathcal{E}(x),\alpha,\epsilon,\epsilon',t}[\lVert\epsilon-\epsilon_{\theta}(\zmat_t,t,\mathcal{P})\rVert_2^2+\lVert\epsilon'-\epsilon_{\theta}(\zmat'_t,t,\mathcal{P}_{\alpha})\rVert_2^2],
\end{equation}
where $\epsilon$ and $\epsilon'$ are random Gaussian noises, and $\zmat_t$ and $\zmat'_t$ are the respective noisy latent codes obtained from $\zmat_0$. 
To preserve the rich image prior learned by the DM, we restrict the number of fine-tuning steps to a small value, typically around 150 steps.

\subsection{Age Editing: Image Inversion}
\label{sec:inversion} After the specialization stage, our DM can generate face images either unconditionally or conditionally on a target age $\alpha$ with prompts $\mathcal{P}$ and $\mathcal{P}_{\alpha}$ respectively. To enable real image editing, we need to inverse the diffusion process of the input image. In this task, we leverage an inversion algorithm, known as \textit{null-text inversion}~\cite{mokady2022null}, which consists in modifying the unconditional textual embedding that is used for classifier-free guidance such that it leads to accurate reconstruction. To be specific, we use the specialized model to invert the input image $\xmat$ to the noise space through DDIM inversion \cite{songdenoising}. We obtain a diffusion trajectory $\{z^{inv}_t \}, t=1\dots T$ from Gaussian noise to the input image. 
Unfortunately, previous studies~\cite{songdenoising} show that classifier-free guidance amplifies the accumulated error of DDIM inversion, resulting in poor reconstruction of $\xmat$.
\textit{Null-text inversion} optimizes the null-text embedding $\varnothing_t$ used in classifier guidance at every step $t$ such that, assuming a conditioning prompt $\mathcal{P}_{inv}$ corresponding to the input image, the forward process leads to an accurate reconstruction of $\xmat$.
The unconditionally inverted sequence of noisy latents $\{z_t^{inv}\}_{t=1}^T$ serves as our pivot trajectory for optimization: the unconditional null embeddings over all time-steps $\{\varnothing_t\}_{t=1}^T$ are sequentially optimized such that the noise estimator network $\epsilon_\theta$ predicts latent codes close to $z_{t-1}^{inv}$ at every step $t$.
More precisely, for every step $t$ in the order of the diffusion process $t = T \rightarrow t = 1$, the following minimization problems are sequentially considered:
\begin{equation}
    \underset{\varnothing_t}\min\lVert z_{t-1}^{inv}-z_{t-1}(\bar{z_t},t,\mathcal{P}_{inv};\varnothing_t)\rVert_2^2
\end{equation}
where $\bar{z_t}$ is the noisy latent code obtained by solving the optimization problem of the previous step, and $z_{t-1}$ is the latent code at step $t-1$ estimated using $\bar{z_t}$.
To enable age editing,  we need to provide a prompt corresponding to the content of the input image. 
In this task, we propose to employ a pre-trained age estimator. 
Assuming an input image $\xmat$, we obtain its estimated age $\alpha$ and employ as prompt $\mathcal{P}_{inv}=\mathcal{P}_{\alpha}=$"\textit{photo of a $[\alpha]$ year old person}".

\subsection{Age Editing: Localized Age Editing with Attention Control}
\label{sec:editing}

We now explain how we edit an image $\xmat$ to make the person in the image appear to be of a target age $\alpha_{\tau}$. To achieve this, we take inspiration from recent literature~\cite{hertz2022prompt} and act on the cross-attention maps used for text-conditioning, forcing the model to modify only age-related areas via attention map injection. After inversion, we know the latent noise $\zmat_T$ and the optimized unconditional embeddings $\{\varnothing_t\}_{t=1}^T$ leads to an accurate reconstruction of $\xmat$ when conditioned on prompt $\mathcal{P}_{\alpha}$. In every cross-attention layer of $\epsilon_\theta$, we compute the reference cross-attention maps generated during the diffusion process $\{M_t^{\alpha}=\text{Softmax}(Q^{\zmat}_t K^{\alpha}_t)\}_{t=1}^T$, where $Q^{\zmat}_t$ are queries computed from $\zmat_t$ and $ K_t^{\alpha}$ keys computed from the prompt $\mathcal{P}_{\alpha}$. As shown in ~\cite{tang2022daam, hertz2022prompt}, these attention maps contain rich semantic relations between the spatial layout of the image and each word in $\mathcal{P}_{\alpha}$. In our case, the attention maps corresponding to the token $[\alpha]$ indicate which pixels are related to the age of the person.

Next, we replace the initial estimated age ${\alpha}$ in the inversion prompt $\mathcal{P}_{\alpha}$ with a target age $\alpha_{\tau}$ and obtain a new target prompt $\mathcal{P}_{\tau}$="\textit{photo of a $[\alpha_{\tau}]$ year old person}". We then use $\mathcal{P}_{\tau}$ to guide the generation: during the new sampling process, we inject the cross-attention maps $\{M_t^{\alpha}\}_{t=1}^T$, but keep the cross-attention values from the new prompt $\mathcal{P}_{\tau}$. 
In this way, the generated image is conditioned on the target age information provided by the target prompt $\mathcal{P}_{\tau}$ through the cross-attention values, while preserving the original spatial structure. Specifically, as only age-related words are modified in the new prompt, only pixels that attend to age-related tokens receive the greatest attention. Note that, we follow \cite{hertz2022prompt} and perform a soft attention constraint by swapping only the first $t_M$ steps, as the attention maps play an important role mostly in the early stages.

\paragraph{Enhancing prompts}
\label{sec:prompt}
\method~can achieve satisfying aging performance with the very generic prompts given above. Nevertheless, the results can be further improved by using more specific prompts in the inversion and editing stages. While this can be achieved with manual prompt engineering, we propose a simple and automatic way to improve our initial prompts $\mathcal{P}_{\alpha}$ and $\mathcal{P}_{\tau}$.
First, we can leverage pre-trained gender classifiers to predict the gender of the person in the input image. Then, the word \textit{"person"} in both $\mathcal{P}_{\alpha}$ and $\mathcal{P}_{\tau}$ can be replaced by either \textit{"woman"} or \textit{"man"}. Second, our experiments show that in the case of young ages, either in $\mathcal{P}_{\alpha}$ and $\mathcal{P}_{\tau}$, the use of words such as \textit{"person"}, \textit{"woman"} or \textit{"man"} do not perform well. Therefore, if the target age $\alpha_{\tau}$ or the age $\alpha$ estimated by our classifier is below 15, the words \textit{"woman/man"} are replaced by \textit{"girl/boy"} in  $\mathcal{P}_{\tau}$ or $\mathcal{P}_{\alpha}$.



\section{Experiments}
\paragraph{Implementation details}
We employ Stable Diffusion \cite{rombach2022high} pre-trained on the LAION-400M dataset \cite{schuhmann2021laion}. 
150 training images are sampled from FFHQ-Aging\cite{or2020lifespan} to finetune the pre-trained model for 150 steps, with a batch size of 2. We used the central age of the true label age group as $\alpha$ in the finetuning prompt $\mathcal{P}_{\alpha}$. 
We employed Adam optimizer with a learning rate of $5\times10^{-6}$ and $\beta_1=0.9$, $\beta_2=0.999$. 
During attention control, we set the cross-attention replacing ratio $t_M/T$ to 0.8.
All experiments are conducted on a single A100 GPU. It takes 1 minute for finetuning, 1 minute for inversion, and 5 seconds for age editing.

\paragraph{Evaluation protocol}
We utilized two widely-used high-resolution \textbf{datasets} as in \cite{gomez2022custom}. \textit{FFHQ-Aging}~\cite{or2020lifespan} is an extension of the NVIDIA FFHQ~\cite{karras2019style} dataset containing 70k 1024$\times$1024 resolution images. Images are manually labeled into 10 age groups ranging from 0-2 to 70+ years old. 
\textit{CelebA-HQ}~\cite{karras2017progressive} consists of 30k images.  This dataset is used only for evaluation, not for training.
Age labels are obtained using the DEX classifier~\cite{rothe2015dex} as used in previous studies~\cite{yao2021high,zhang2017age}. 
Images are downsampled to 512$\times$512 resolution for our experiments.
Regarding the \textbf{metrics}, 
we evaluate aging methods from three perspectives: aging accuracy, age-irrelevant attribute reservation, and aging quality. Following \cite{gomez2022custom}, we employ:
\textit{Mean Absolute Error (MAE)}: the prediction of an age estimator is compared with the target age.
\textit{Gender, Smile, and Face expression preservation}: we report the percentage to which the original attribute is preserved.
\textit{Blurriness}: indicates face blur condition.
\textit{Kernel-Inception Distance} \cite{binkowski2018demystifying} assesses the discrepancy between generated and real images for similar ages. 
We report the KID between original and generated images within the same age groups.
For evaluation, Face++\footnote{Face++ Face detection API: https://www.faceplusplus.com/} is used for aging accuracy, attribute preservation, and blurriness evaluation.

\subsection{Comparison with State-of-the-Art}
We conduct comparisons with state-of-the-art aging approaches, including HRFAE~\cite{yao2021high}, LATS~\cite{or2020lifespan}, and CUSP~\cite{gomez2022custom}. We are unable to include Re-aging GAN~\cite{makhmudkhujaev2021re}, another recent aging method, in our comparison due to the unavailability of its source code. Moreover, the lack of detailed information regarding its evaluation protocol prevents us from conducting a fair and reliable comparison following its evaluation protocol.
We start the comparison on the CelebA-HQ\cite{karras2017progressive} dataset. In this case, we follow the evaluation protocol used in \cite{yao2021high} and sample 1000 test images with \textit{"young"} labels and translate them to the target age of 60. 

\begin{figure}[t]
  \centering
  \begin{tabular}{cc}
    \includegraphics[width=0.92\textwidth]{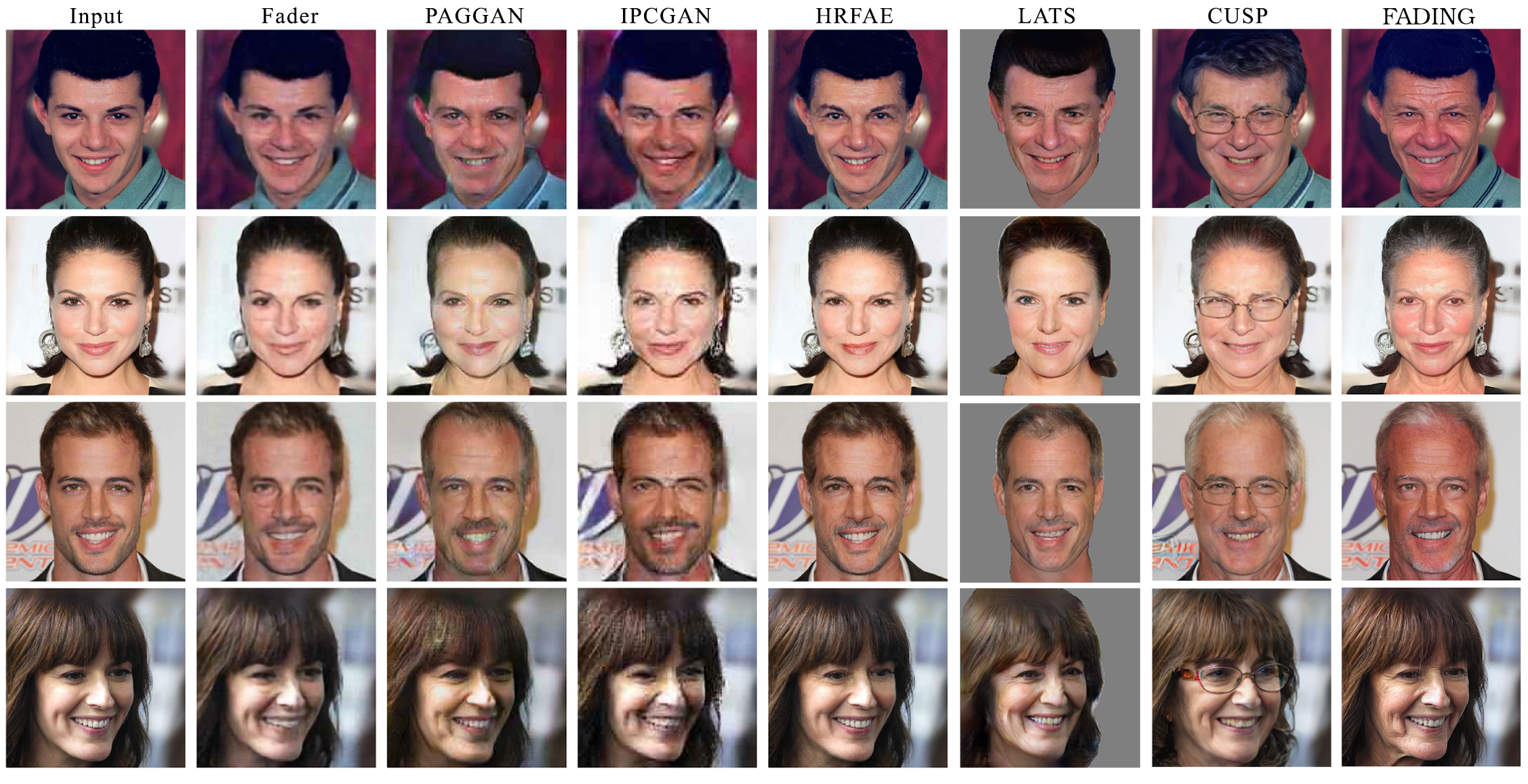} 
      \end{tabular}
  \caption{Qualitative comparison with state-of-the-art methods on CelebA-HQ. Images for the other approaches are extracted from \cite{gomez2022custom}.}
  \label{fig:celeba_res}
\end{figure}

\paragraph{Qualitative comparison}
The comparative study on CelebA-HQ is shown in Figure~\ref{fig:celeba_res}. Note that these images are extracted from ~\cite{gomez2022custom}, and consequently have not been cherry-picked. We observe that FaderNet~\cite{lample2017fader} introduces little modifications, PAG-GAN~\cite{yang2019learning} and IPC-GAN~\cite{wang2018face} produce pronounced artifacts or degradation. HRFAE~\cite{yao2021high} generates plausible aged faces with minor artifacts but is mostly limited to skin texture changes, such as adding wrinkles. LATS~\cite{or2020lifespan}, CUSP~\cite{gomez2022custom}, and our approach introduce high-level semantic changes, such as significant receding of the hairline (see third row). But LATS operates only in the foreground; it does not deal with backgrounds or clothing and requires a previous masking procedure. On the other hand, CUSP always introduces glasses with aging. This is likely due to the high correlation between age and glasses in their training set. Our method does not introduce these undesired additional accessories, produces fewer artifacts on backgrounds, and possesses more visual fidelity to the input image.

We now expand the comparison with the best-performing competitor, namely CUSP~\cite{gomez2022custom}, on FFHQ-Aging~\cite{or2020lifespan}. We translate input images to all age groups and report per-age-group results, for a more comprehensive analysis with a complete sense of continuous transformation throughout the lifespan. Figure~\ref{fig:ffhq} shows qualitative results.
We have the following key observations.
(1) In general, our approach introduces fewer artifacts, generates realistic textural and semantic modification, and achieves better visual fidelity across all age groups.
(2) We achieve significant improvement for extreme target ages (infant and elderly, see columns for (4-6) and (70+)). 
(3) Our model handles better rare cases, such as accessories or occlusions. CUSP fails when the source person wears facial accessories. Typically, for the person on the right who wears sunglasses, CUSP falsely translates sunglasses to distorted facial components. In contrast, our method preserves accessories accurately while correctly addressing structural changes elsewhere. These results confirm our initial hypothesis that utilizing a specialized DM pre-trained on a large-scale dataset increases robustness compared to methods exclusively trained on facial datasets, which are susceptible to data bias.

Interestingly, we observe a slight variation in skin tone  when  addressing age change with~\method. It is important to  note that a similar shift in skin tone is also observed for the  training-free baseline (vanilla implementation of prompt-to-prompt editing using pretrained Stable Diffusion, referred  to as Training-free in Table~\ref{tab:abla:spec}), as shown in Figure~\ref{fig:abla_EP} (see more results in supplementary material). This suggests that the entanglement between age and skin tone is inherent to the pre-trained Stable Diffusion model and is not a result of  our specialization stage.

\begin{figure}[t]
 \def\myim#1{ \includegraphics[width=12.0mm,height=12.0mm]{#1}}
\centering
   \setlength\tabcolsep{0.5 pt}
   \renewcommand{\arraystretch}{0.2}

     \begin{tabular}{lccccccccccccc}
     
        &\scriptsize Input &\scriptsize (4-6) &\scriptsize (20-29) &\scriptsize (40-49) &\scriptsize (70+) & \hspace{2mm}  &\scriptsize Input &\scriptsize (4-6) &\scriptsize (20-29) &\scriptsize (40-49) &\scriptsize (70+)\\
        
        \rotatebox{90}{\scriptsize CUSP} &
        \myim{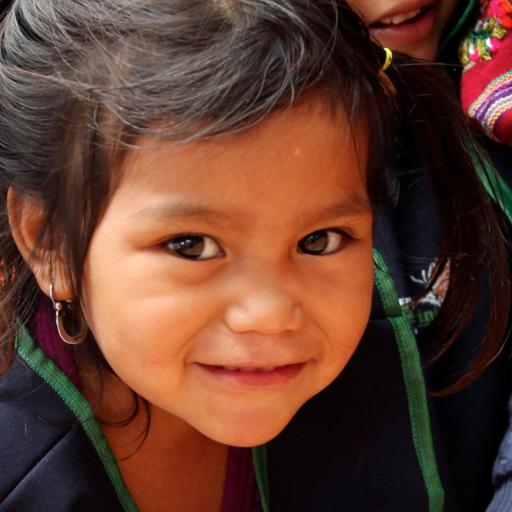} &
        \myim{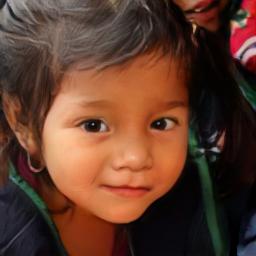} &
        \myim{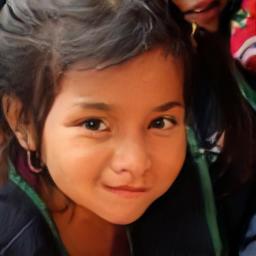} &
        \myim{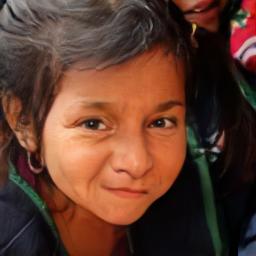} &
        \myim{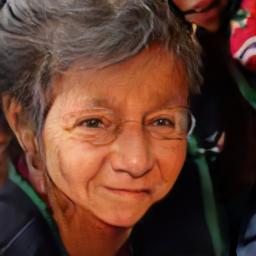} &
        &
        \myim{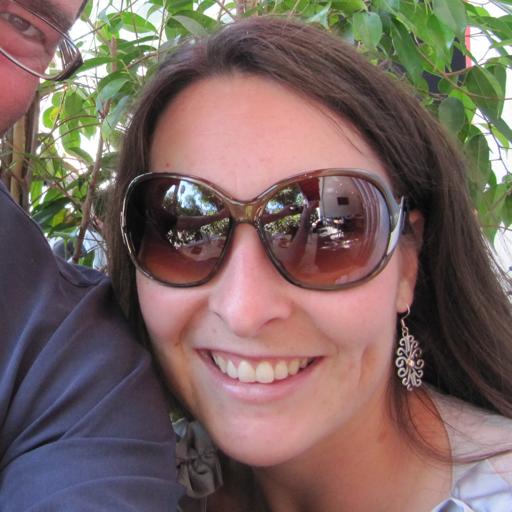} &
        \myim{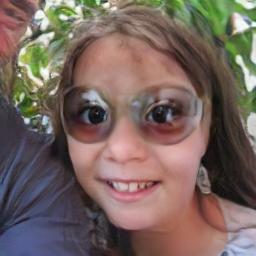} &
        \myim{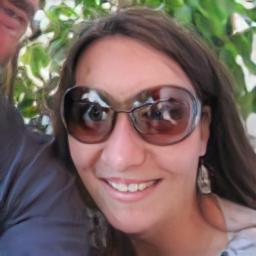} &
        \myim{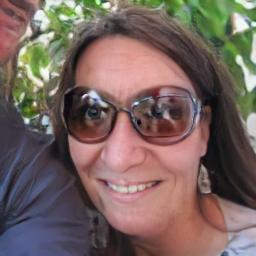} &
        \myim{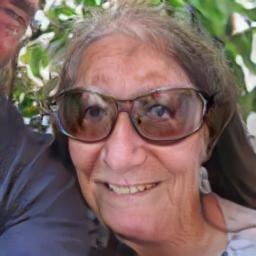} \\
        
        \rotatebox{90}{\scriptsize \method} &
        \myim{images/FFHQ_res/input_00007.jpg} &
        \myim{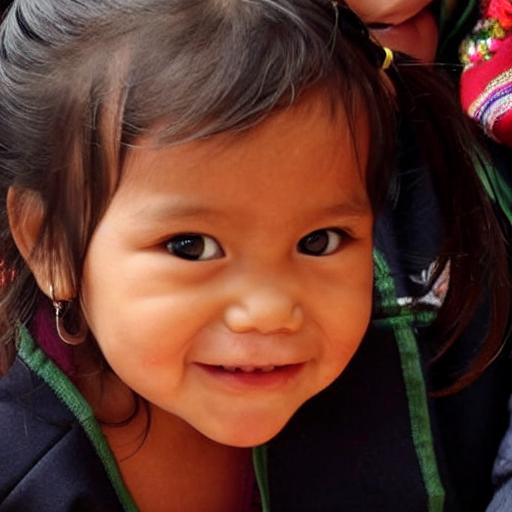} &
        \myim{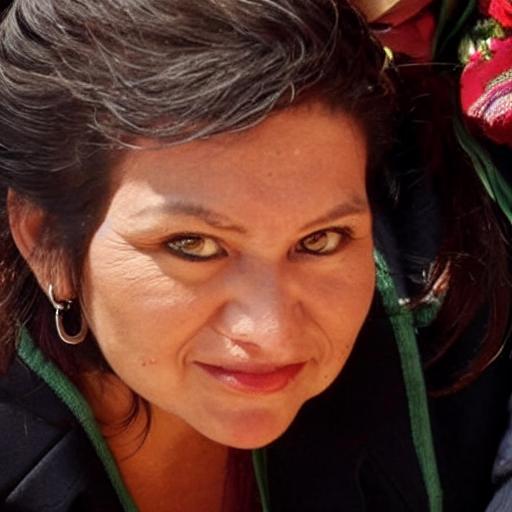} &
        \myim{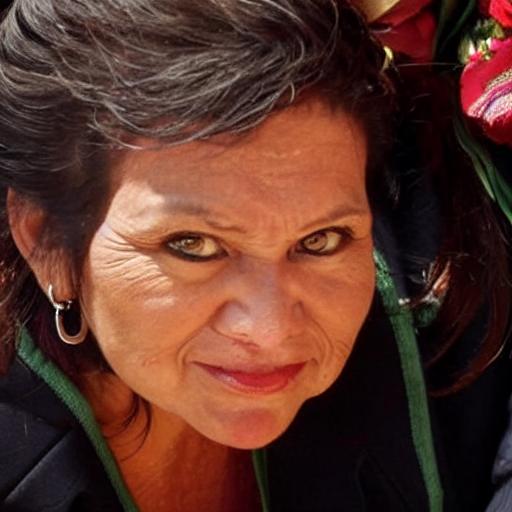} &
        \myim{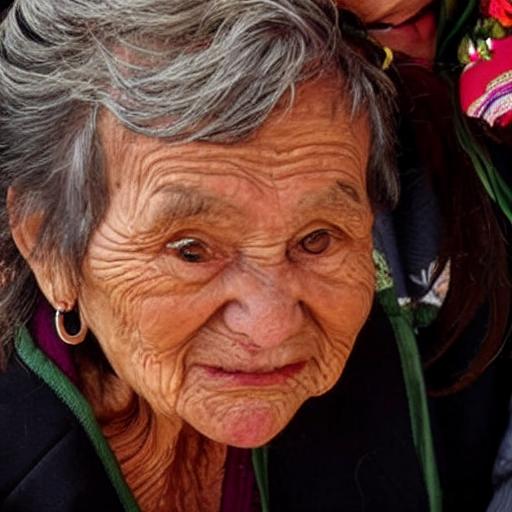} &  
        &
        \myim{images/FFHQ_res/input_00022.jpg} &
        \myim{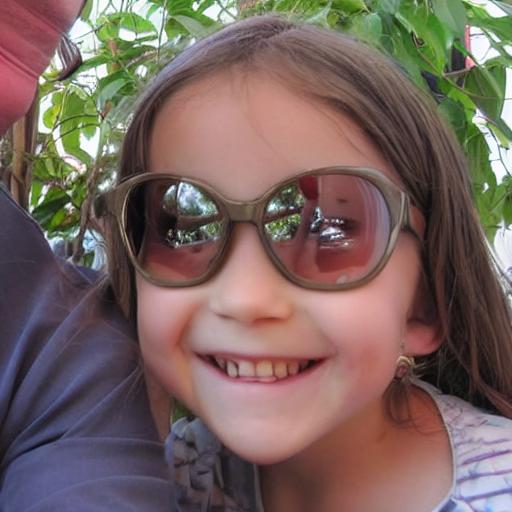} &
        \myim{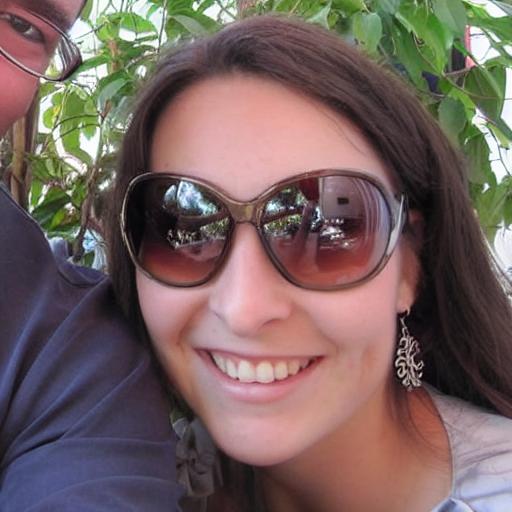} &
        \myim{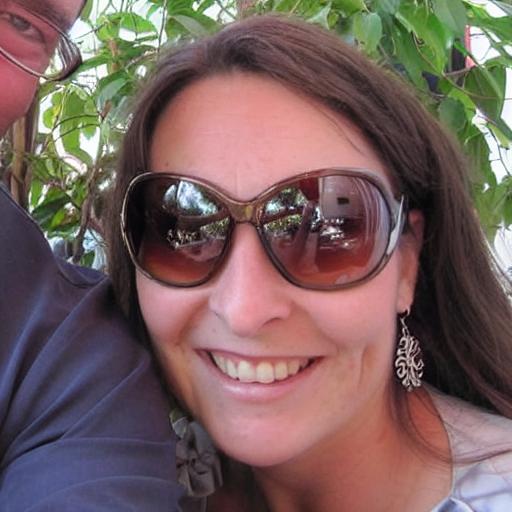} &
        \myim{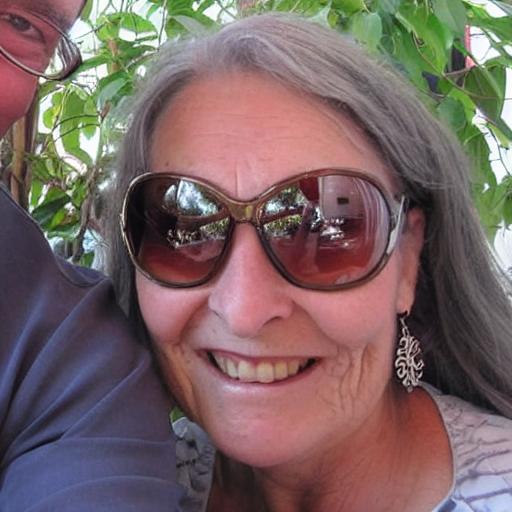} \\

        

    \end{tabular}

\caption{Qualitative comparison with state-of-the-art methods on FFHQ-Aging. For CUSP, we translate each image to the corresponding age group. For \method, we translate to the central age of each group. For the oldest age group (70+), we translate to 80 years old.}
\label{fig:ffhq}
\end{figure}

\paragraph{Quantitative comparison}

\begin{table}[t]
\centering
\caption{Quantitative comparison on CelebA-HQ on the young-to-60 task. Except for \method, the scores are extracted from~\cite{gomez2022custom}.}
\label{tab:celeba}
\fontsize{8}{10}\selectfont
\scriptsize
\begin{tabular}{lcccccc}
\toprule
Method & Predicted Age & Blur & Gender & Smiling & Neutral & Happy \\ \midrule
Real images & 68.23 $\pm$ 6.54 & 2.40 & - & - & - & - \\ \midrule
FaderNet~\cite{lample2017fader} & 44.34 $\pm$ 11.40 & 9.15 & 97.60 & 95.20 & 90.60 & 92.40 \\ 
PAGGAN~\cite{yang2019learning} & 49.07 $\pm$ 11.22 & 3.68 & 95.10 & 93.10 & 90.20 & 91.70 \\
IPCGAN~\cite{wang2018face} & 49.72 $\pm$ 10.95 & 9.73 & 96.70 & 93.60 & 89.50 & 91.10 \\
HRFAE~\cite{yao2021high} & 54.77 $\pm$ 8.40 & \textbf{2.15} & 97.10 & \textbf{96.30} & \textbf{91.30} & \textbf{92.70} \\
HRFAE-224~\cite{yao2021high} & 51.87 $\pm$ 9.59 & 5.49 & 97.30 & 95.50 & 88.30 & 92.50 \\
LATS~\cite{or2020lifespan} & 55.33 $\pm$ 9.33 & 4.77 & 96.55 & 92.70 & 83.77 & 88.64 \\
CUSP~\cite{gomez2022custom} & \textbf{67.76 $\pm$ 5.38} & 2.53 & 93.20 & 88.70 & 79.80 & 84.60 \\
\method~(Ours) & 66.49 $\pm$ 6.46 & 2.35 & \textbf{98.40} & 90.20 & 84.50 & 86.80 \\ \bottomrule
\end{tabular}
\end{table}

Table~\ref{tab:celeba} presents quantitative results on CelebA-HQ\cite{karras2017progressive} dataset. Note that an 8.23-year discrepancy is reported between the DEX classifier utilized for inference and the Face++ classifier utilized for evaluation\cite{gomez2022custom}. \method~is on par with CUSP for aging accuracy. We achieve the highest gender preservation, proving our capability to retain age-irrelevant features. However, we report lower scores for other attributes. As is discussed in the qualitative analysis, this is because previous methods primarily generate texture-level modification, which preserves high-level attributes. In contrast, \method~yields more profound but realistic semantic changes, thus slightly compromising preservation metrics. 

\begin{table}[t]
    \centering
    \caption{Quantitative comparison between CUSP and \method~on FFHQ-Aging.}
    \label{tab:ffhq}
       \resizebox{0.9\columnwidth}{!}{ 
       \begin{tabular}{llccccccccccc}
      \toprule
        Metric & Method & 0-2 & 3-6 & 7-9 & 10-14 & 15-19 & 20-29 & 30-39 & 40-49 & 50-69 & 70+ & Mean \\
      \midrule
        \multirow{2}{*}{MAE} & CUSP & 9.41 & 16.28 & 20.24 & 18.16 & 11.88 & 10.36 & 12.70 & \textbf{11.08} & \textbf{8.13} & 8.05 & 12.63 \\
         & \method & \textbf{5.70} & \textbf{11.72} & \textbf{13.66} & \textbf{11.22} & \textbf{6.86} & \textbf{6.23} & \textbf{9.60} & 12.04 & 8.39 & \textbf{6.20} & \textbf{9.16} \\
         \midrule
        \multirow{2}{*}{Gender(\%)} & CUSP & 71.5 & \textbf{73.5} & \textbf{74.5} & \textbf{78.0} & 73.5 & 80.5 & 85.5 & 81.5 & 82.0 & 76.0 & 77.7 \\
         & \method & \textbf{72.0} & 72.0 & 67.5 & 68.0 & \textbf{88.0} & \textbf{96.0} & \textbf{98.0} & \textbf{97.0} & \textbf{95.0} & \textbf{87.5} & \textbf{84.1} \\
               \midrule
        \multirow{2}{*}{KID($\times$100)} & CUSP & 4.19 & 3.22 & 3.14 & 3.18 & 3.60 & 3.63 & 3.98 & 4.69 & 4.07 & 4.57 & 3.83 \\
         & \method & \textbf{1.41} & \textbf{0.11} & \textbf{0.45} & \textbf{0.25} & \textbf{0.52} & \textbf{0.16} & \textbf{1.00} & \textbf{0.59} & \textbf{1.50} & \textbf{0.61} & \textbf{0.66} \\
              \bottomrule
            \end{tabular}}
\end{table}

Table~\ref{tab:ffhq} presents quantitative results on FFHQ-Aging\cite{or2020lifespan} dataset. Lower MAE suggests that we have a better aging accuracy. \method~also reports better gender preservation for most age groups. Note that, for middle-aged group from 30-50, an almost perfect preservation rate is achieved. Our qualitative analysis is supported by the quantitative KID analysis, with one order of magnitude lower than CUSP for nearly all age groups. Again this demonstrates that \method~achieves higher aging performance.

\begin{figure*}[t]
     \centering
      \begin{subfigure}[b]{0.32\textwidth}
         \centering
         \includegraphics[height=2.1cm]{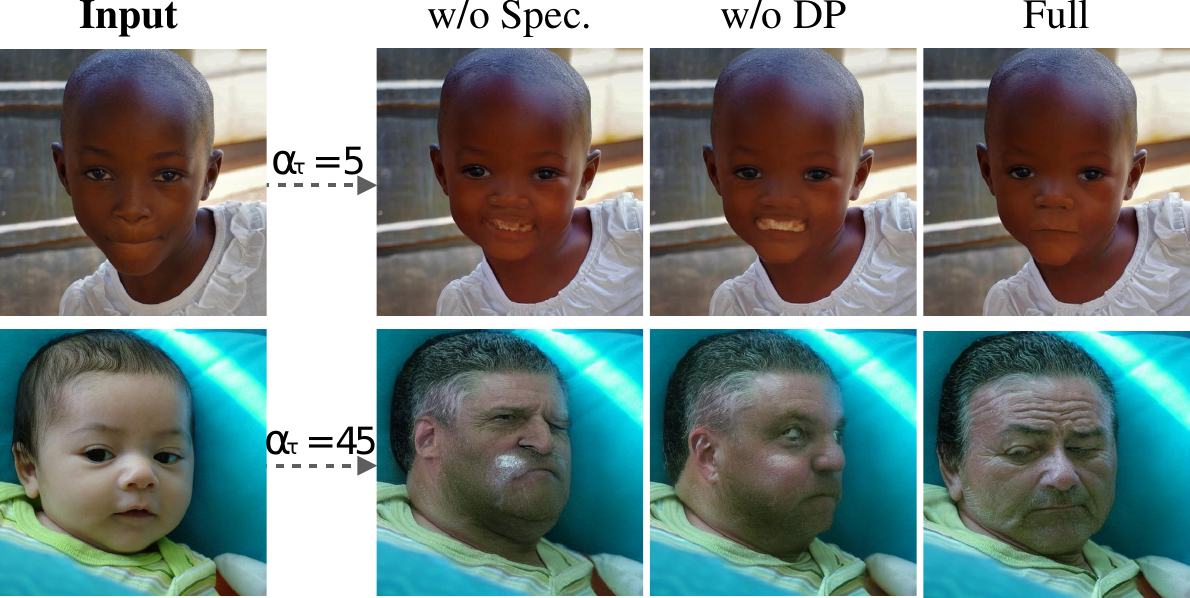}
         \caption{Specialization step}
         \label{fig:abla_spec}
     \end{subfigure}
     \hfill
     \begin{subfigure}[b]{0.32\textwidth}
         \centering
         \includegraphics[height=2.2cm]{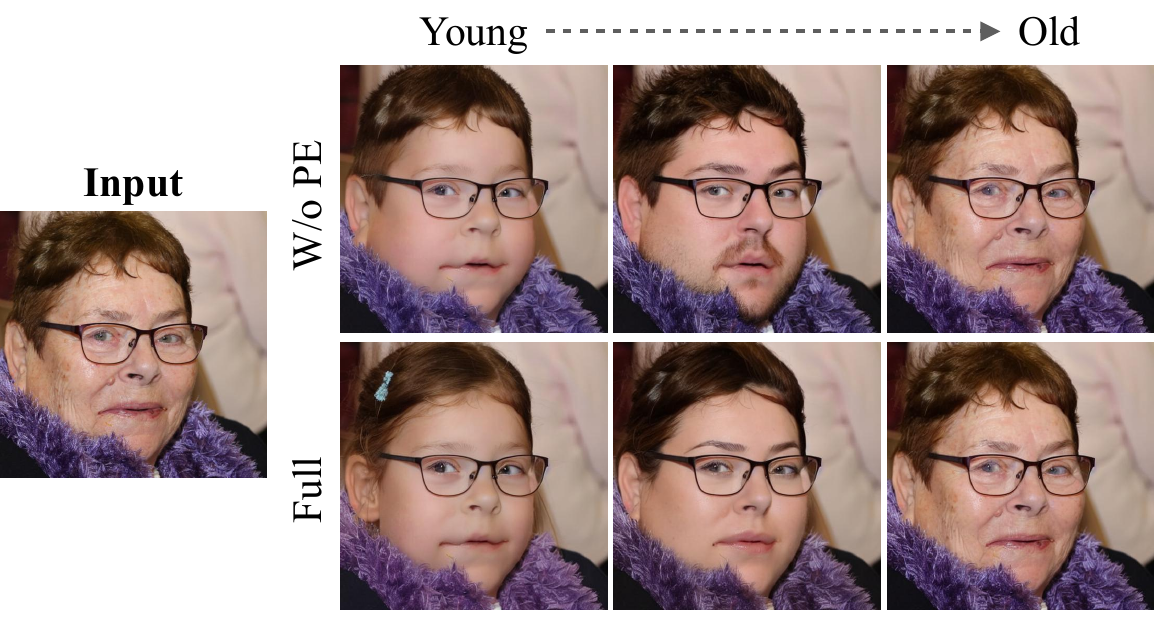}
         \caption{Enhanced Prompts}
         \label{fig:abla_EP}
     \end{subfigure}
     \hfill
     \begin{subfigure}[b]{0.32\textwidth}
         \centering
         \includegraphics[height=2.2cm]{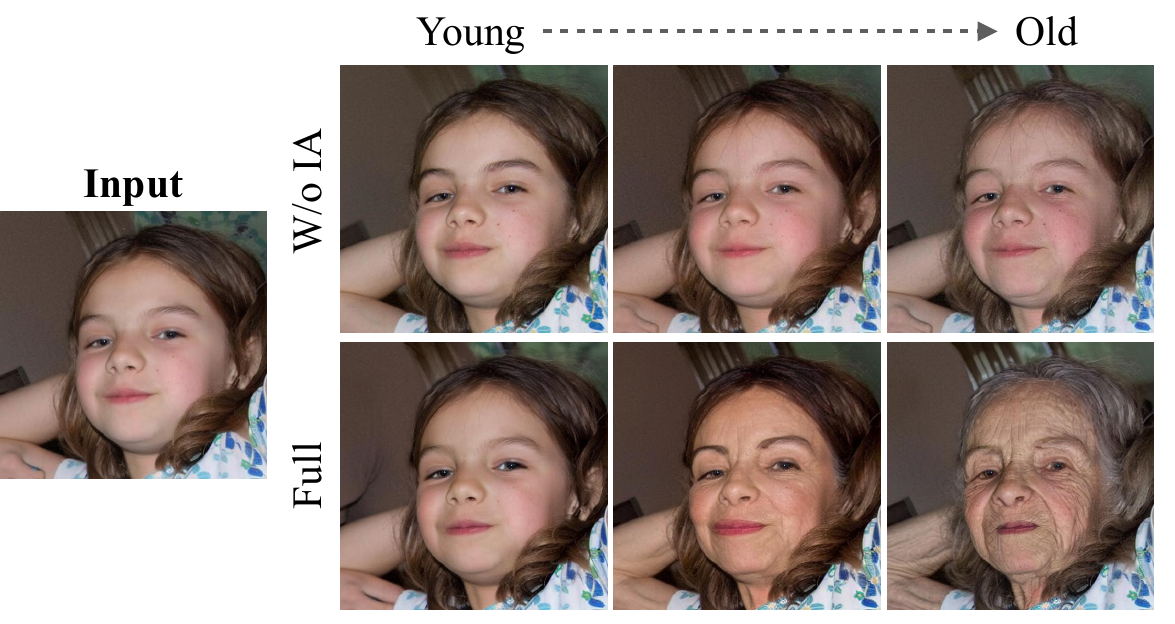}
         \caption{Initial Age}
         \label{fig:abla_IA}
     \end{subfigure}
            \caption{Qualitative ablation studies on several aspects of \method: impact of the specialization step (\textit{Spec.}), the use of Double Prompt (\textit{DP}) , the Enhanced Prompts (\textit{EP}) and the use of the Initial Age (\textit{IA}).
        }
        \label{fig:abla}
        \vspace{-4mm}
\end{figure*}

\subsection{Ablation studies}



\begin{table}[t]
  \begin{minipage}[b]{.68\linewidth}
    \centering
    \caption{Ablation study on the Specialization stage}
\label{tab:abla:spec}
\scriptsize \setlength{\tabcolsep}{1.1pt}
    \begin{tabular}{lcccccccccc}
    \toprule
    Method & Spec. & DP &   MAE & Gender & Smiling & Happy & Neutral & Blur & KID{\tiny($\times$100)}\\
    \midrule
     Training-free~$\sim$\cite{mokady2022null}  & \xmark & - & 9.295 & 82.40 & 82.95 & 78.35 & 78.80 & 2.226 & 0.668 \\
     Single prompt & \cmark & \xmark & \textbf{8.781} & 81.95 & 85.05 & 81.55 & 81.05 & 2.275 & 0.707 \\

     Full & \cmark & \cmark &9.162 & \textbf{84.10} & \textbf{86.60} & \textbf{81.95} & \textbf{81.75} & \textbf{2.030} & \textbf{0.660}\\
    \bottomrule
    \end{tabular}
  \end{minipage}%
  \hfill
    \begin{minipage}[b]{.3\linewidth}
    \centering
\caption{Ablation study on the Editing stage}
\label{tab:ablaEditing}
\scriptsize \setlength{\tabcolsep}{1.1pt}
    \begin{tabular}{lccc}
    \toprule
    Method &    MAE & Gender & KID{\tiny($\times$100)} \\
    \midrule
     w/o EP  &9.830 & 79.90 & 0.668 \\
    w/o IA & 13.703 & 80.05 & 1.164\\
     Full &   \textbf{9.162} & \textbf{84.10} &\textbf{ 0.660}\\
    \bottomrule
    \end{tabular}
  \end{minipage}
\end{table}

\paragraph{Specialization (Spec.) and Double-Prompt (DP) scheme}
 To assess the influence of the design of the specialization step, we consider a variant where we skip the specialization step and directly use a pre-trained Stable Diffusion instead. This baseline can be seen as a vanilla implementation of prompt-to-prompt editing \cite{hertz2022prompt} with null-text inversion \cite{mokady2022null} in the case of aging. The second variant includes the specialization step but omits the double-prompt scheme.
The results shown in Figure~\ref{fig:abla_spec} and Table~\ref{tab:abla:spec} demonstrate the effectiveness of our specialization step in generating more realistic images.
Our qualitative analysis indicates that the images edited with a non-specialized model exhibit noticeable aberrations, especially around the mouth area and facial contours. The quantitative metrics also support the observation that our method achieves higher aging quality (lowest blurriness and KID). Furthermore, the training-free editing approach reports the highest aging error and a low attribute preservation rate. 
Regarding our double-prompt scheme, Figure~\ref{fig:abla_spec} shows that it improves the structural alignment with the original image. Quantitatively, as shown in Table~\ref{tab:abla:spec}, the slight increase in age-MAE brought by \textit{DP} is vastly complemented by the large gains in attribute preservation metrics. 
This improvement suggests that the \textit{DP} indeed enhances the disentanglement of age from age irrelevant features by keeping them better retained. Besides, the age-MAE metric may be a less strong indicator of disentanglement capability, given that differences of 0.38 year in facial appearance are often imperceptible in real photos.

\paragraph{Enhanced Prompts (EP) and Initial Age (IA)}
We now analyze the edition stage considering two other variants: one without our enhanced prompts and another which does not use the initial age of the source image and instead uses $(\mathcal{P},\mathcal{P_\tau})$ as editing prompts.
The positive impacts of enhanced prompts and the use of the estimated initial age are demonstrated in Table~\ref{tab:ablaEditing} where we observe consistent gains in all metrics.
Qualitatively, \textit{EP} plays an important role in preserving age irrelevant attributes: we observe significant improvements in gender consistency in Figure~\ref{fig:abla_EP}.
Surprisingly, the use of gender information in our enhanced prompts also helps to improve aging accuracy. We hypothesize that this is because more detailed prompts (we assume that "woman" contains more information than "person") lead to more specialized  attention maps for each semantic component, resulting in more accurate targeting of age-related pixels. The impact of \textit{IA} is illustrated in Figure ~\ref{fig:abla_IA}. Without information on the initial age, the appearance of the person barely changes, except for slight variations in hair color. This indicates that the use of initial age (\textit{IA}) in guiding  prompts prevents the model from reproducing the original image without effectively addressing the age change.


\section{Conclusion}

In this paper, a novel method for face age editing based on diffusion  models was presented. The proposed model leverages the rich image and semantic prior of large-scale text-image models, via a training stage that specializes the diffusion model for aging tasks.  Qualitative and quantitative analyses on two different datasets  demonstrated that our method produces natural-looking re-aged faces  across a wider range of age groups with higher re-aging accuracy, better aging quality, and greater robustness compared to state-of-the-art methods. The effectiveness of each component of our method was also  validated  through extensive experiments. In future works, we plan to extend our enhanced prompts strategy to preserve other age-agnostic attributes by leveraging corresponding pre-trained attribute classifiers. For example, we could include \textit{"wearing glasses"} in the editing prompts when glasses are detected.

\section*{Acknowledgments}

This paper has been supported by the French National Research Agency (ANR-
20-CE23-0027).

\bibliography{main}

\pagebreak
\appendix
\section*{Appendix}

\noindent
In this supplementary material, we report additional qualitative examples. 

\section{Additional Comparisons with CUSP~\cite{gomez2022custom}}

In  Figure~\ref{fig:ffhq}, we compare with CUSP~\cite{gomez2022custom} on the FFHQ-aging dataset for images of people with different initial ages and ethnicities. In line with the observation discussed in the main paper, \method~is able to generate images with fewer artifacts and better adherence to the original identity across the lifespan. Specifically, we show the results on rare cases (see the last row for photos of profiles). While CUSP fails with such extreme poses, \method~still manages to generate realistic aging results.

\begin{figure}[h]
 \def\myim#1{ \includegraphics[width=12.0mm,height=12.0mm]{#1}}
\centering
   \setlength\tabcolsep{0.5 pt}
   \renewcommand{\arraystretch}{0.2}

     \begin{tabular}{lccccccccccccc}
     
        &Input &(4-6) &(20-29) &(40-49) &(70+) & \hspace{2mm}  &Input &(4-6) &(20-29) &(40-49) &(70+)\\
        
        \rotatebox{90}{\scriptsize CUSP} &
        \myim{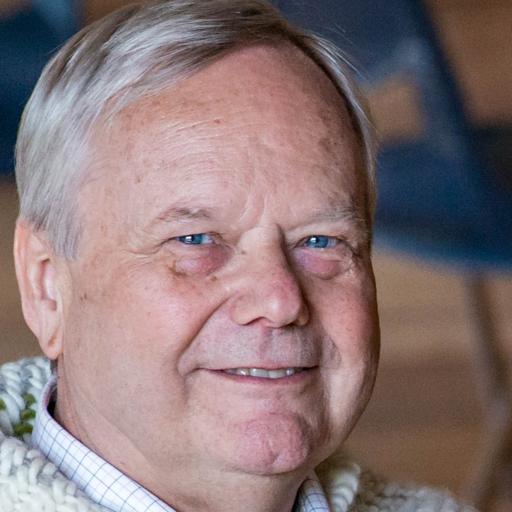} &
        \myim{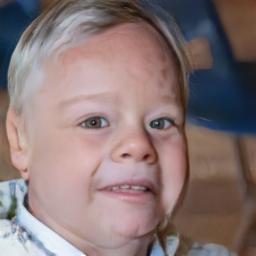} &
        \myim{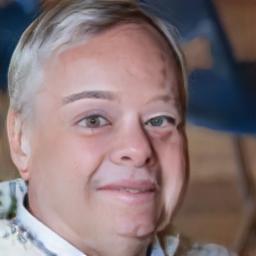} &
        \myim{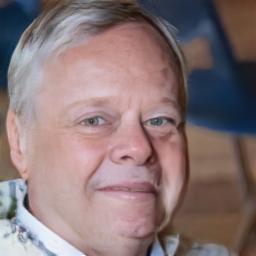} &
        \myim{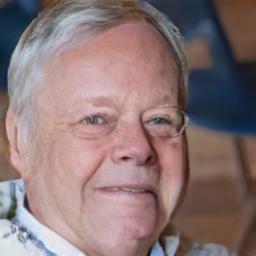} &
        &
        \myim{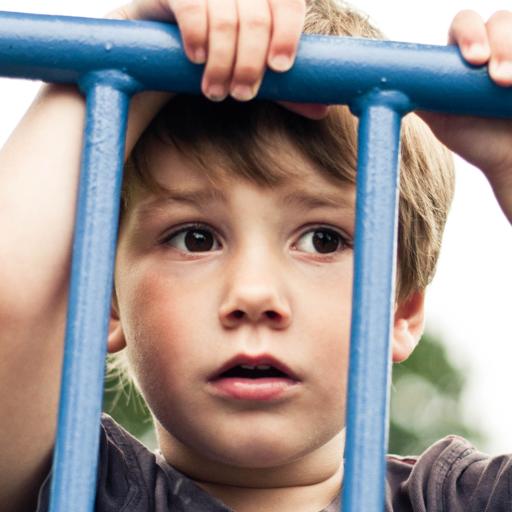} &
        \myim{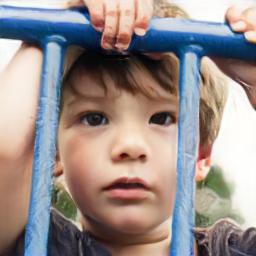} &
        \myim{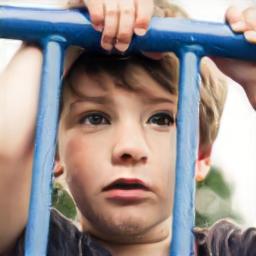} &
        \myim{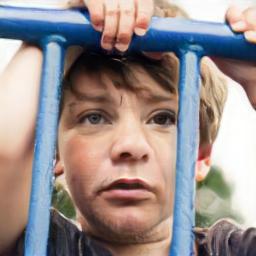} &
        \myim{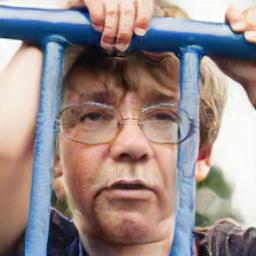} \\
        
        \rotatebox{90}{\scriptsize \method} &
        \myim{images/FFHQ_res/input_00718.jpg} &
        \myim{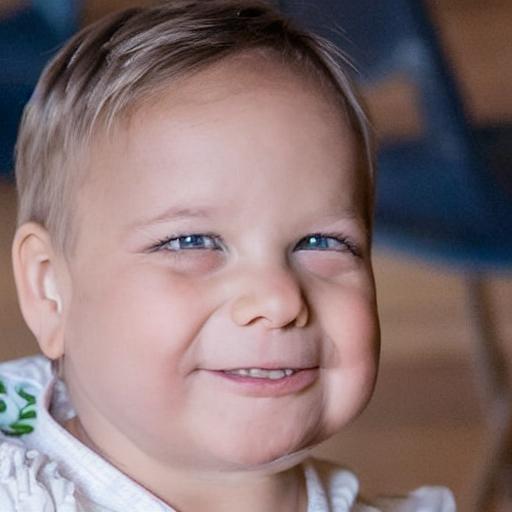} &
        \myim{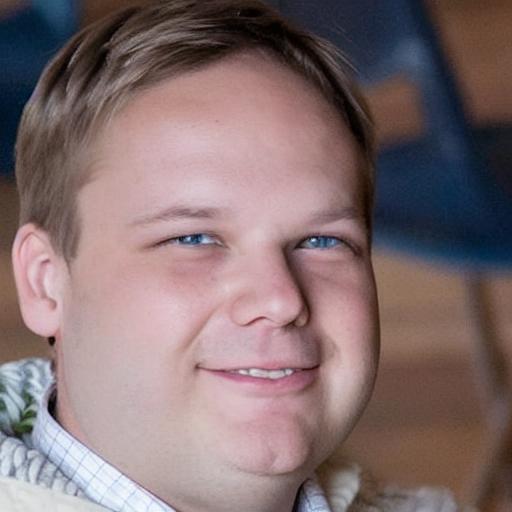} &
        \myim{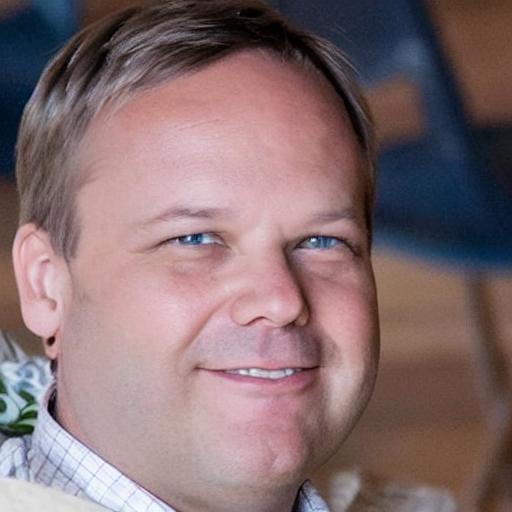} &
        \myim{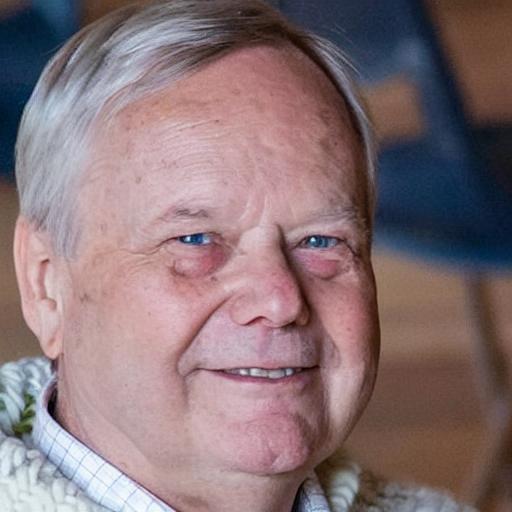} &  
        &
        \myim{images_sup/FFHQ_res/input_00068.jpg} &
        \myim{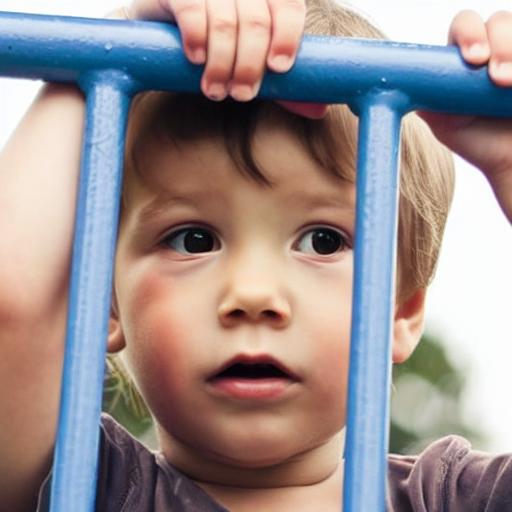} &
        \myim{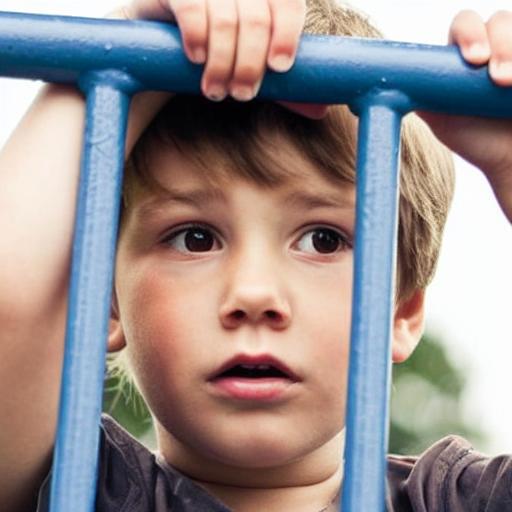} &
        \myim{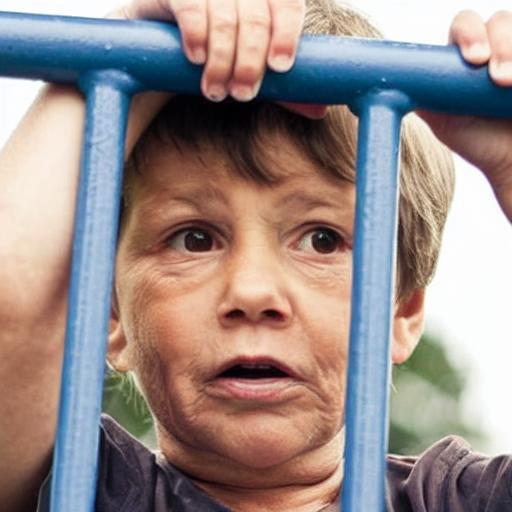} &
        \myim{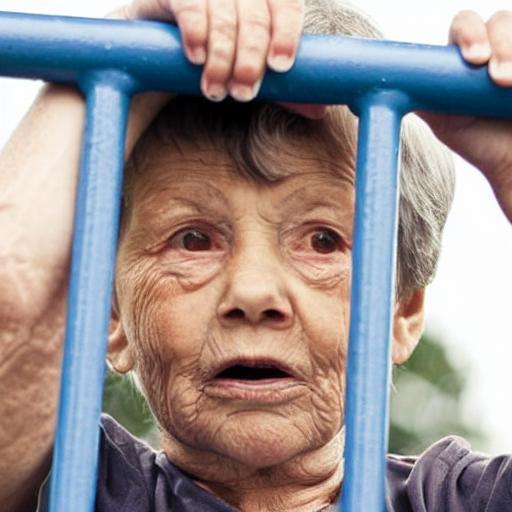} \\

        \rotatebox{90}{\scriptsize CUSP} &
        \myim{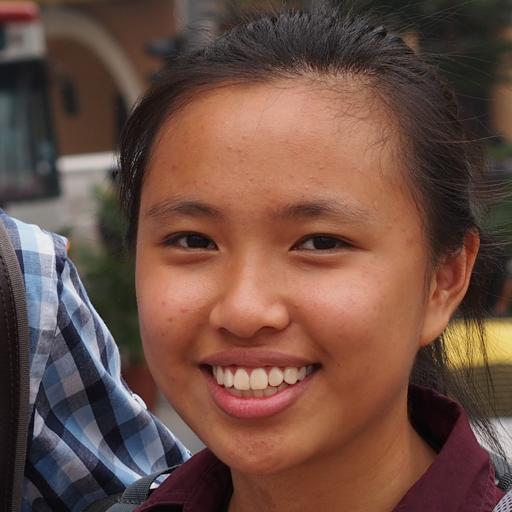} &
        \myim{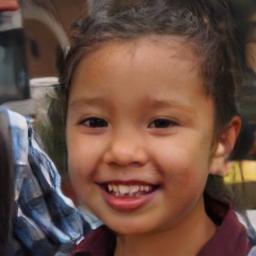} &
        \myim{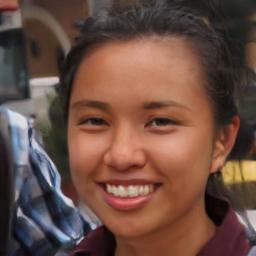} &
        \myim{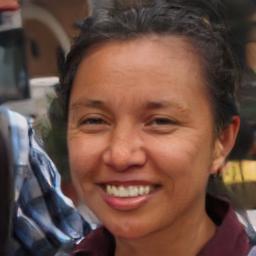} &
        \myim{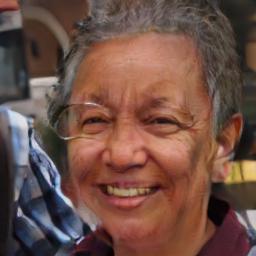} &
        &
        \myim{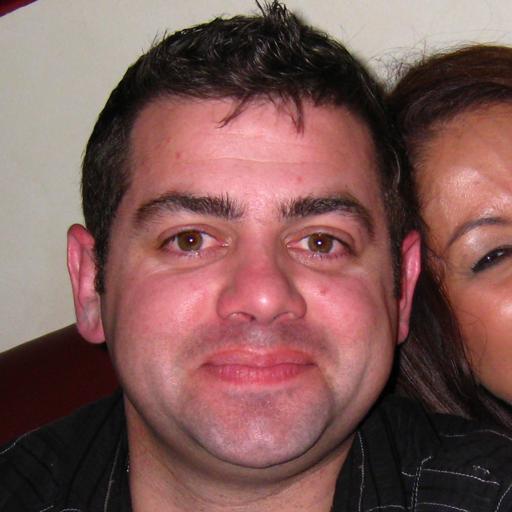} &
        \myim{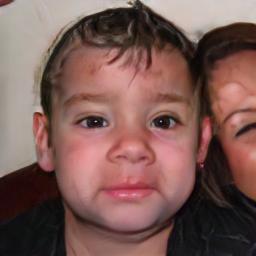} &
        \myim{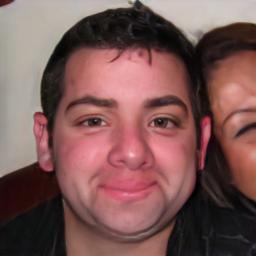} &
        \myim{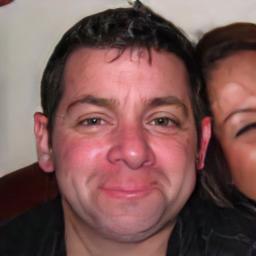} &
        \myim{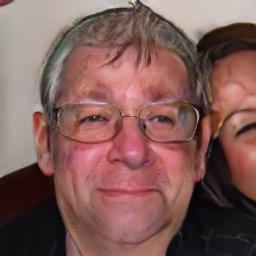} \\
        
        \rotatebox{90}{\scriptsize\method} &
        \myim{images_sup/FFHQ_res/input_00140.jpg} &
        \myim{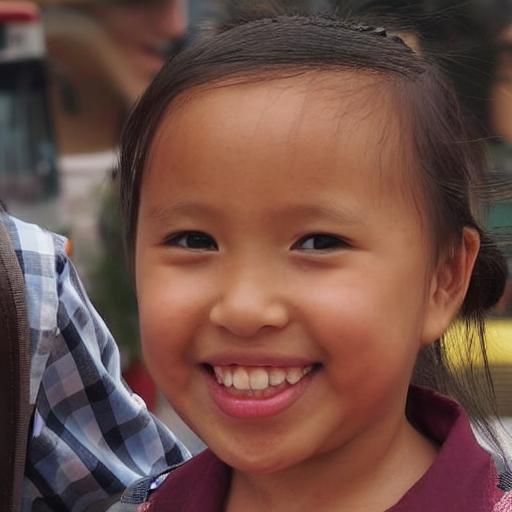} &
        \myim{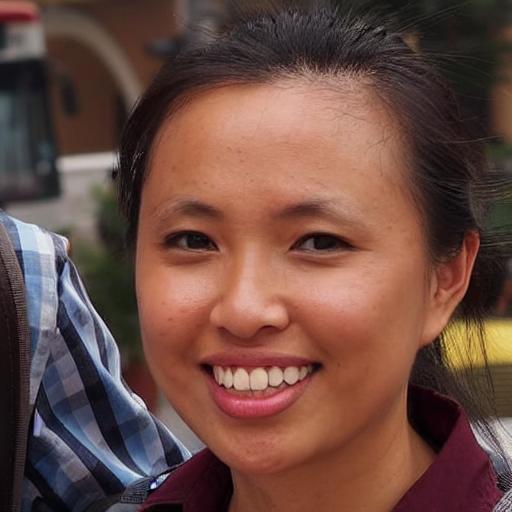} &
        \myim{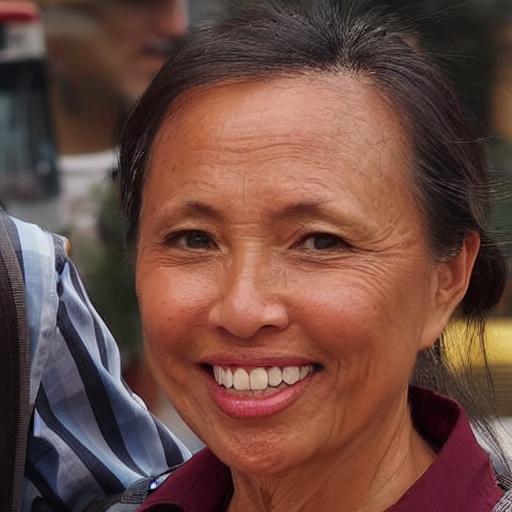} &
        \myim{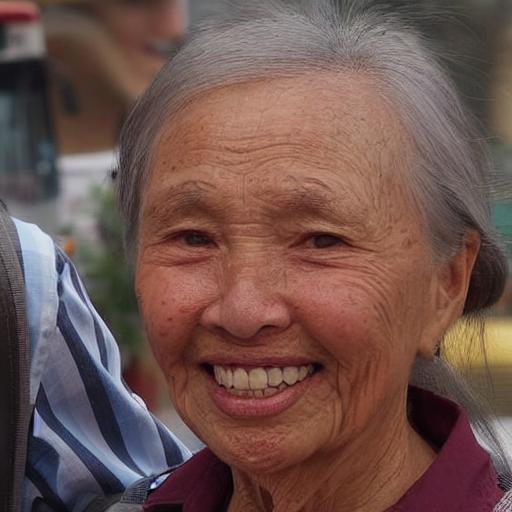} &  
        &
        \myim{images/FFHQ_res/input_00016.jpg} &
        \myim{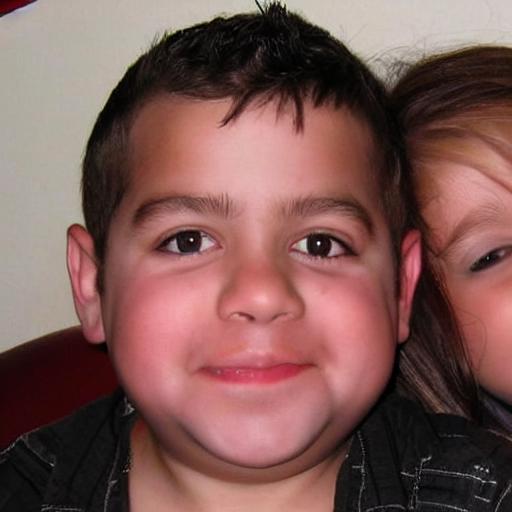} &
        \myim{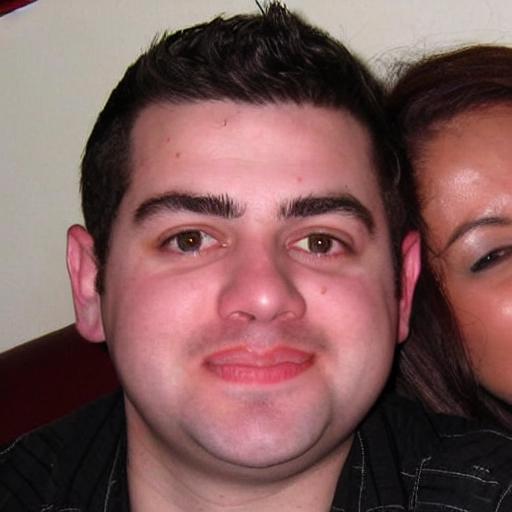} &
        \myim{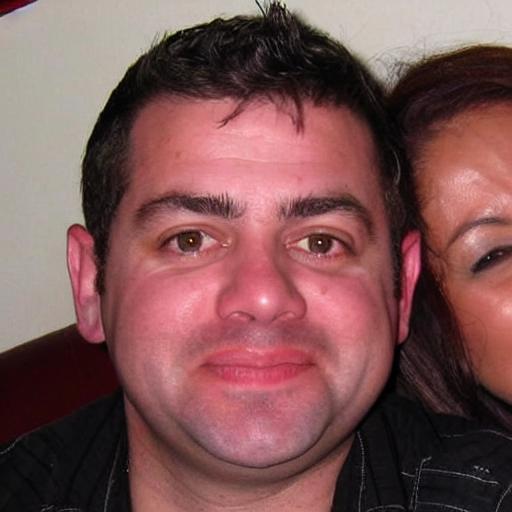} &
        \myim{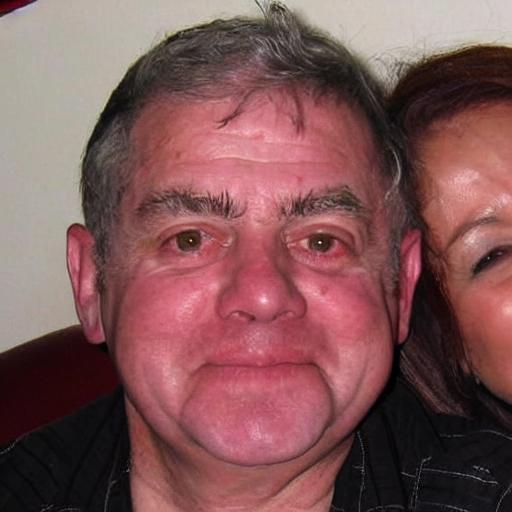} \\   

        \rotatebox{90}{\scriptsize CUSP} &
        \myim{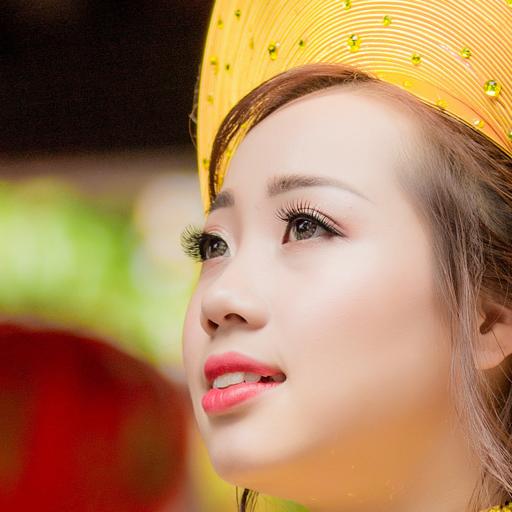} &
        \myim{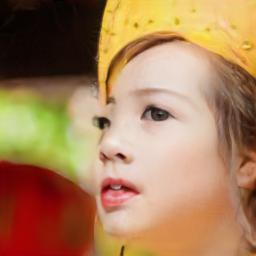} &
        \myim{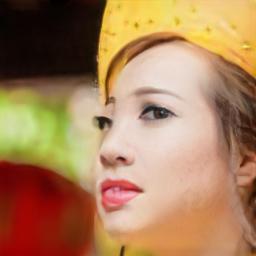} &
        \myim{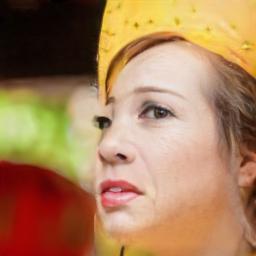} &
        \myim{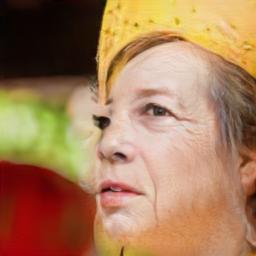} &
        &
        \myim{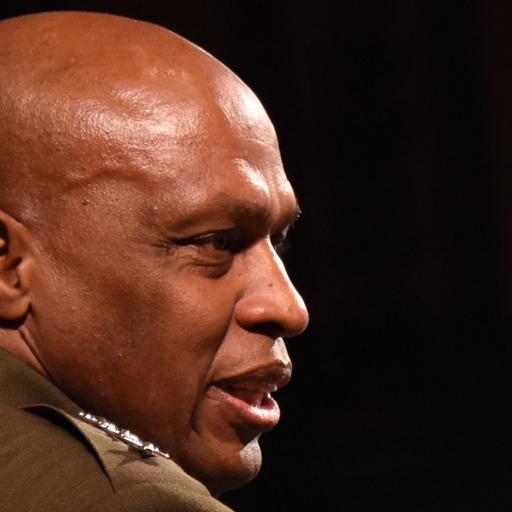} &
        \myim{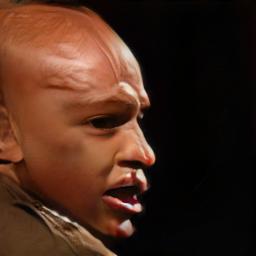} &
        \myim{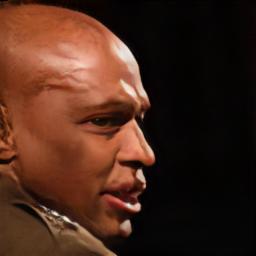} &
        \myim{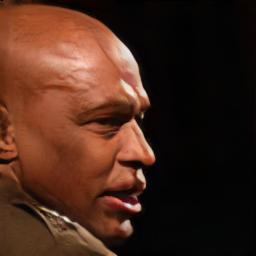} &
        \myim{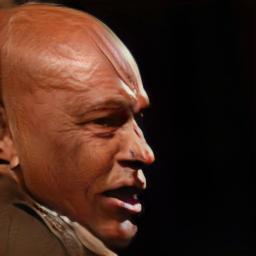} \\
        
        \rotatebox{90}{\scriptsize\method} &
        \myim{images_sup/FFHQ_res/input_00338.jpg} &
        \myim{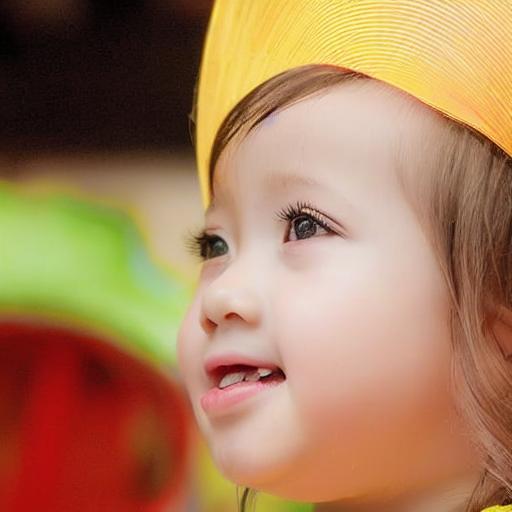} &
        \myim{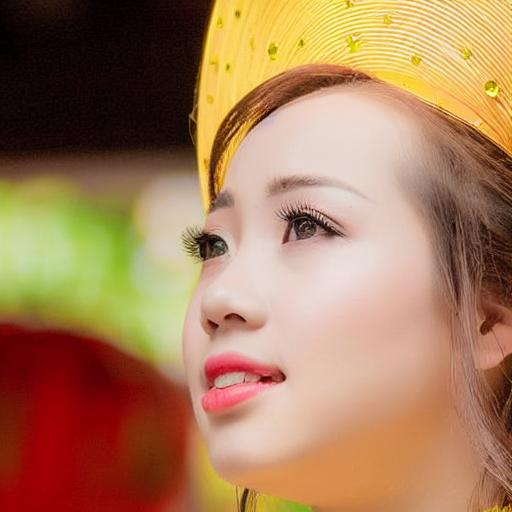} &
        \myim{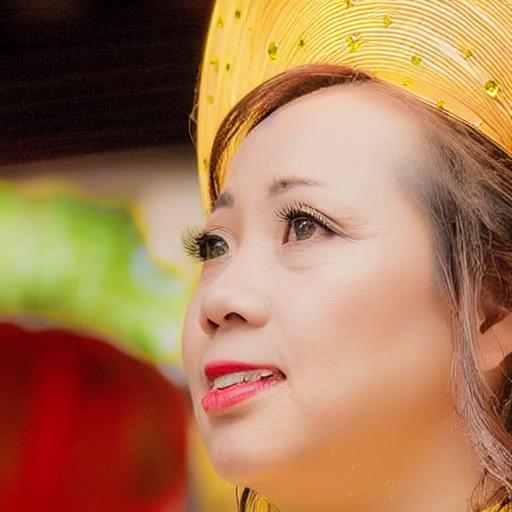} &
        \myim{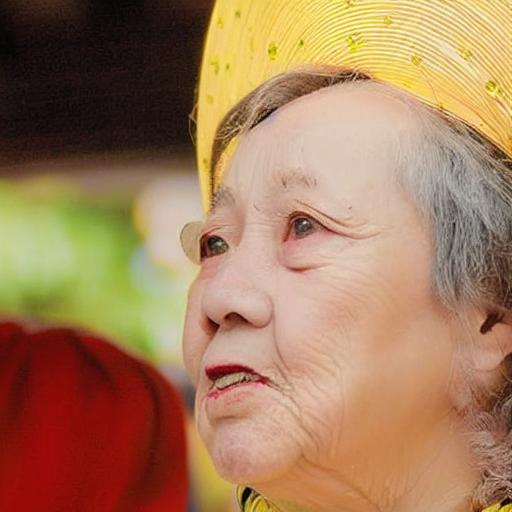} &  
        &
        \myim{images_sup/FFHQ_res/input_00060.jpg} &
        \myim{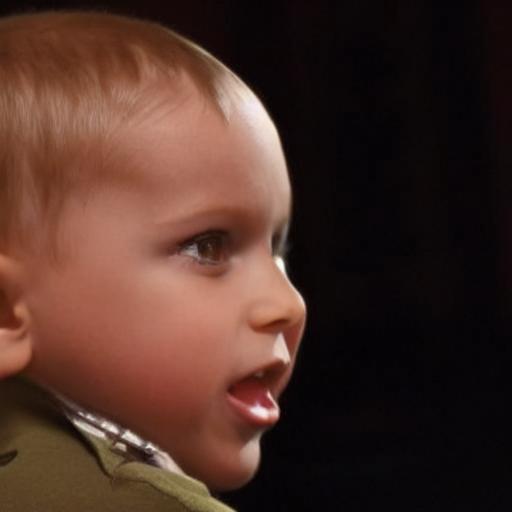} &
        \myim{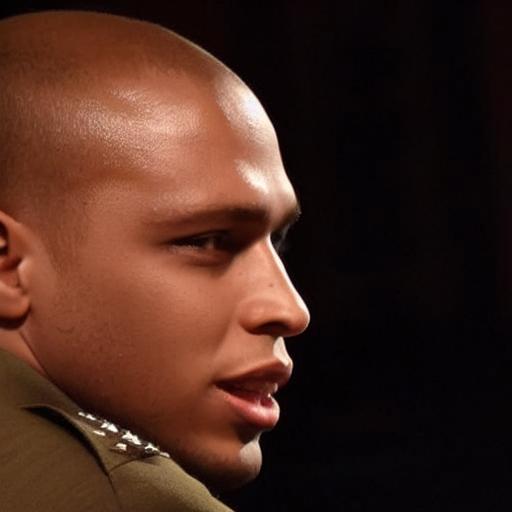} &
        \myim{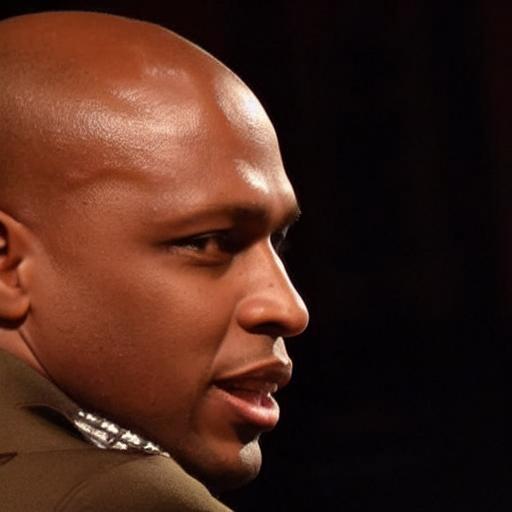} &
        \myim{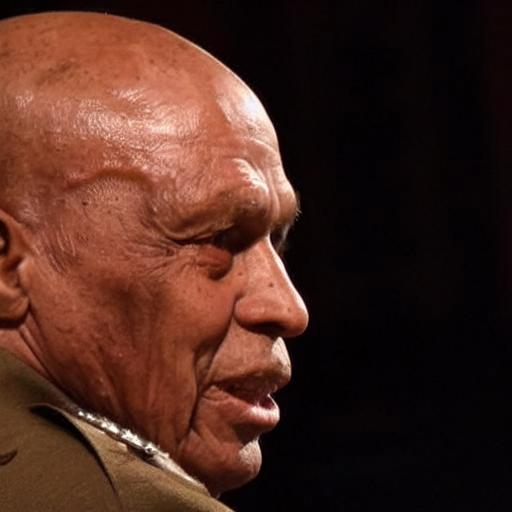} \\

    \end{tabular}

\caption{Supplementary qualitative comparison with CUSP~\cite{gomez2022custom} on FFHQ-Aging. For CUSP, we translate each image to the corresponding age group. For \method, we translate to the central age of each group. For the oldest age group (70+), we translate to 80 years old.}
\label{fig:ffhq}
\end{figure}

\section{Ablation Studies: Additional Examples}
\begin{figure*}[ht]
     \centering
     \begin{subfigure}[b]{0.68\textwidth}
         \centering
         \includegraphics[height=4.8cm]{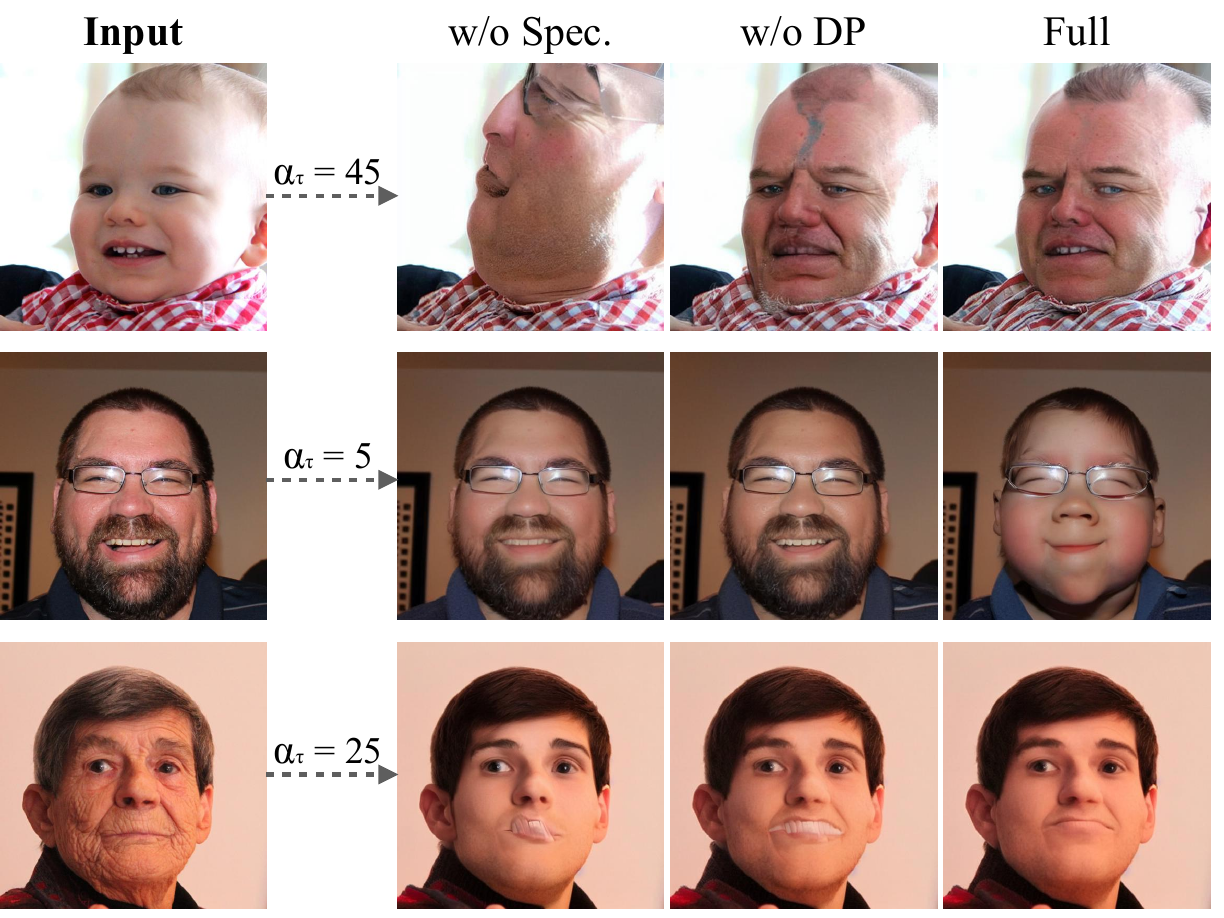}
     \end{subfigure}

    \caption{Supplementary qualitative ablation studies on the impact of the specialization step (\textit{Spec.}) and the use of Double Prompt (\textit{DP}).
    }
    \label{fig:sup_abla_spec}
        \vspace{-4mm}
\end{figure*}

\paragraph{Ablation of \textit{Specialization (Spec.)} and \textit{Double Prompt (DP)}} In Figure~\ref{fig:sup_abla_spec}, we present additional results of ablation experiments performed on the use of \textit{Specialization (Spec.)} and \textit{Double Prompt (DP)}, with different initial ages: young, middle, and old. This further showcases the effectiveness of our specialization step and double-prompt scheme in tackling difficult age-transformation tasks. For example, when the initial age is extremely young (see the first row where the input image is an infant) or when the age gap is big (see the last row where the age difference exceeds 50 years), the specialization step noticeably mitigates the occurrence of artifacts in the generated output. Moreover, we observe that in certain cases where the non-specialized model fails to accurately address age changes (see second row), \textit{Spec.} combined with \textit{DP} contributes to successfully addressing the age alteration.

\begin{figure*}[ht]
     \centering
     \begin{subfigure}[b]{0.48\textwidth}
         \centering
         \includegraphics[height=5.8cm]{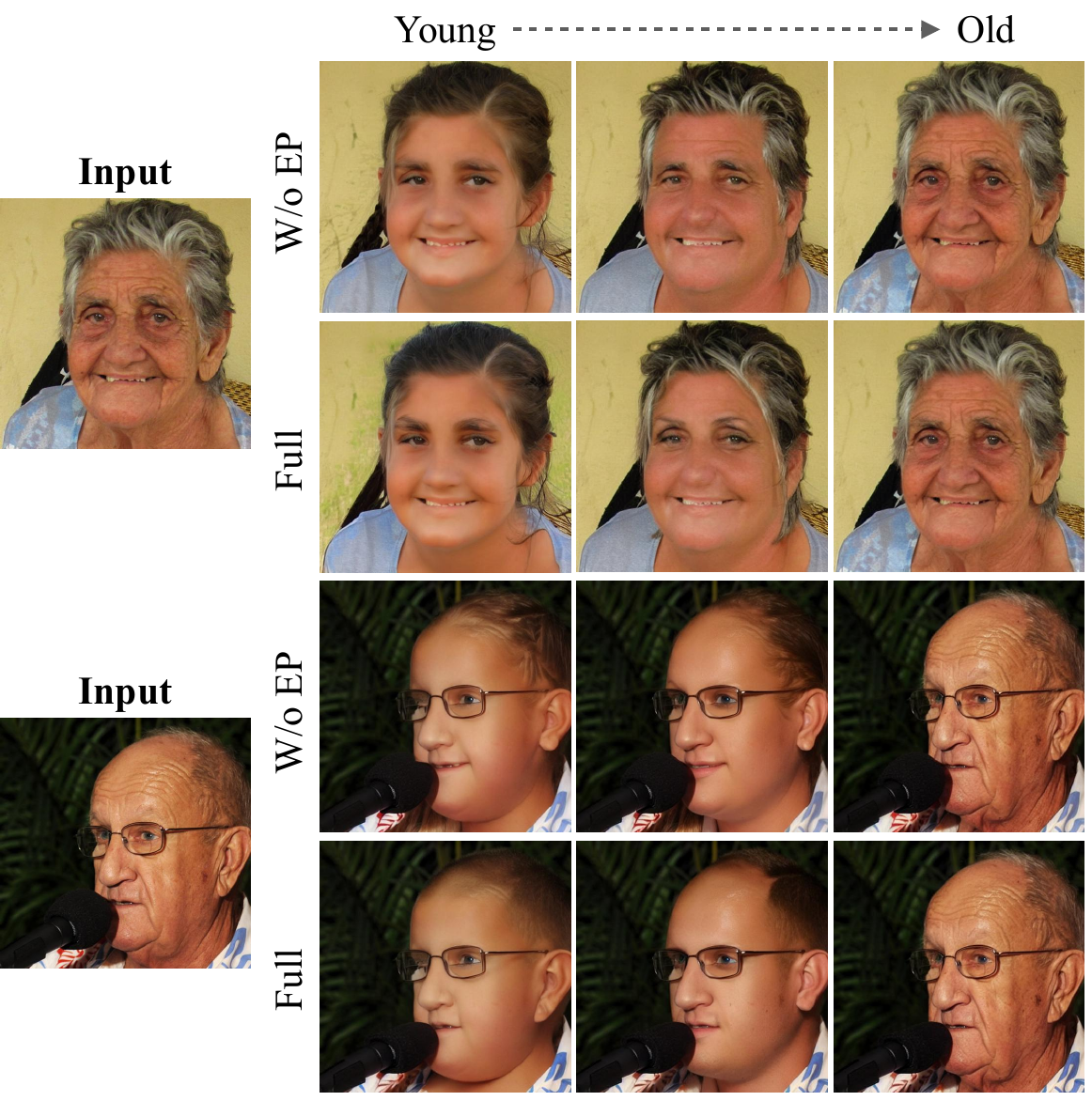}
         \caption{Enhanced Prompts}
         \label{fig:abla_EP}
     \end{subfigure}
\hfill
     \begin{subfigure}[b]{0.48\textwidth}
         \centering
         \includegraphics[height=5.8cm]{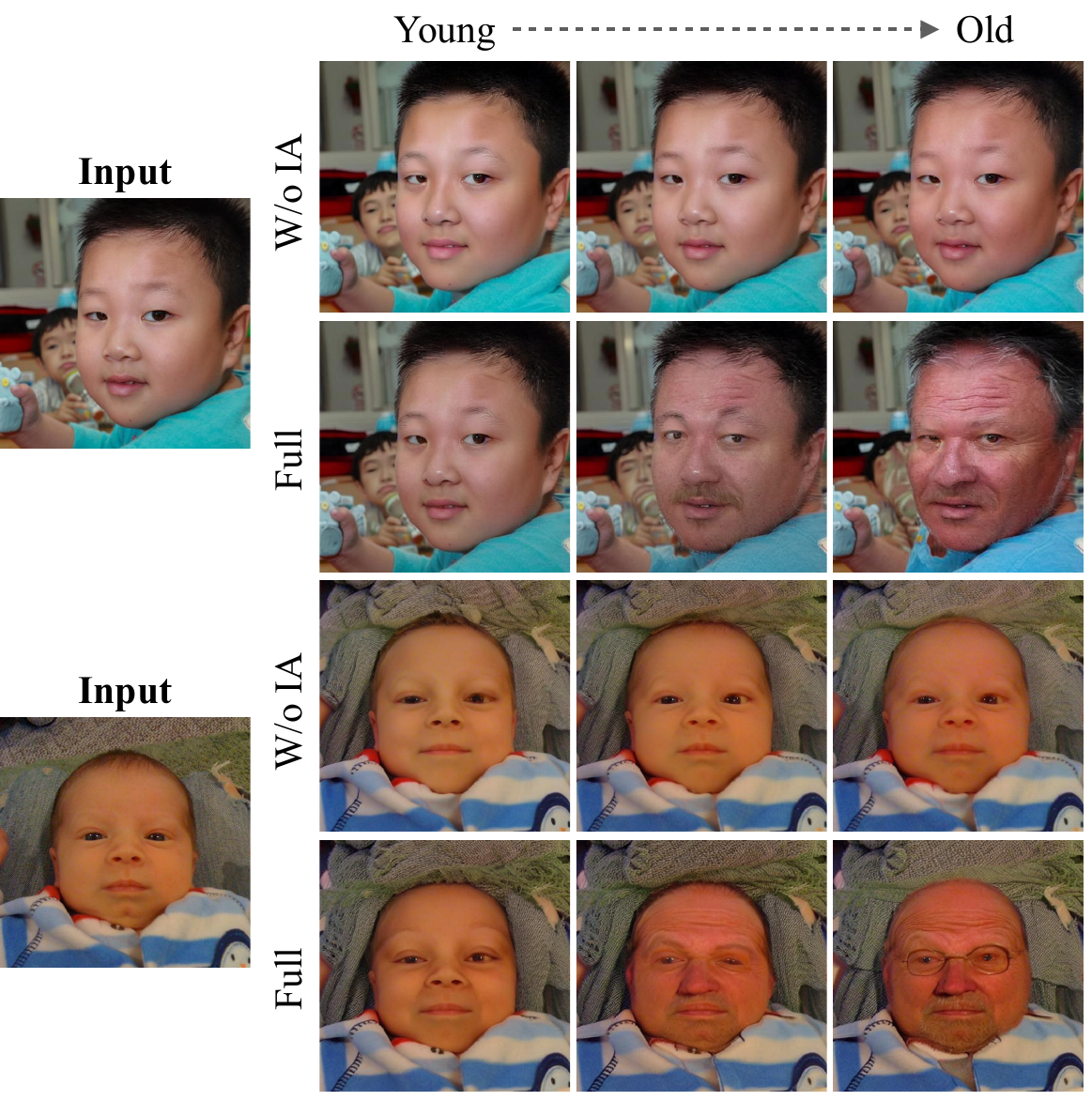}
         \caption{Initial Age}
         \label{fig:abla_IA}
     \end{subfigure}
    \caption{Supplementary qualitative ablation studies on the impact of the Enhanced Prompts (\textit{EP}) and the use of the Initial Age (\textit{IA}).
    }
        \label{fig:sup_abla_EP_IA}
        \vspace{-4mm}
\end{figure*}

\paragraph{Ablation of \textit{Enhanced Prompts (EP)} and \textit{Initial age (IA)}} In Figure~\ref{fig:sup_abla_EP_IA}, we present additional results of ablation experiments conducted on \textit{EP} and \textit{IA}.  When \textit{EP} is removed, we observe that different generated outputs for the same input image exhibit either a different gender compared to the original image (see first example middle column) or gender inconsistencies among the various age-transformed results (see the second example, young and middle-aged results). This underscores the crucial role of \textit{EP} in maintaining gender consistency throughout the age transformation process. On the other hand, when \textit{IA} is ablated, the model struggles to effectively alter the age, particularly when dealing with young initial ages. This indicates that \textit{IA} plays a significant role in facilitating successful age changes, and its removal severely hampers the model's ability to perform accurate age transformations.

\section{Additional Examples on out-of-distribution images} 
\begin{figure}[t]
 \def\myim#1{ \includegraphics[width=13.5mm,height=13.5mm]{#1}}
  \centering

\setlength\tabcolsep{0.5 pt}
   \renewcommand{\arraystretch}{0.22}    
     \begin{tabular}{cccccccc}
        Input &10 &20 &40 &60 & 80  \\
        \myim{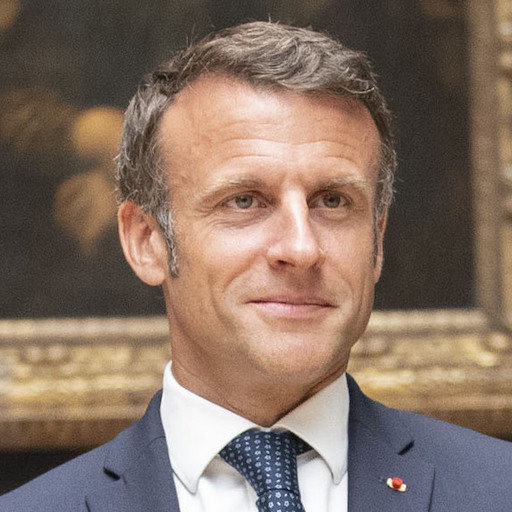} &
        \myim{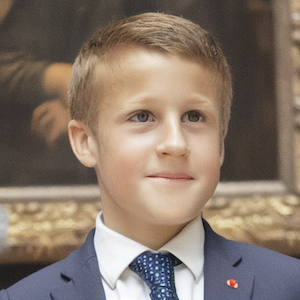} &
        \myim{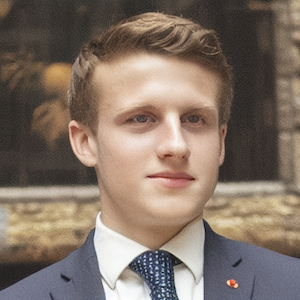} &
        \myim{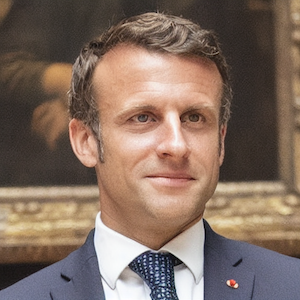} &
        \myim{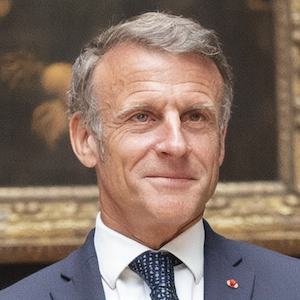} 
        & \myim{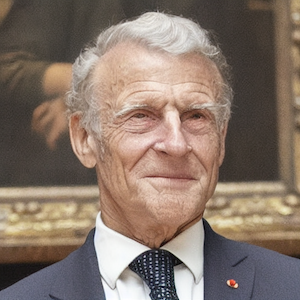} 
         
        \\
        \myim{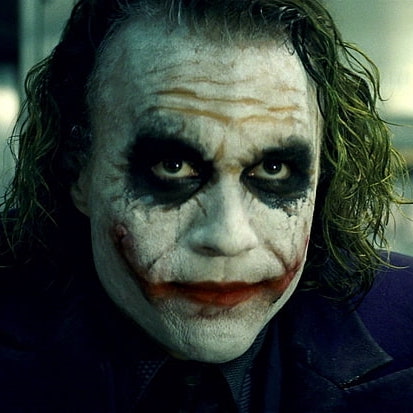} &
        \myim{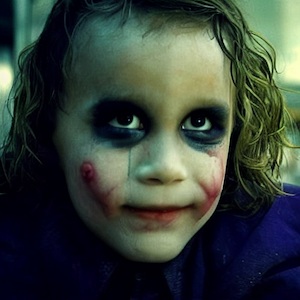} &
        \myim{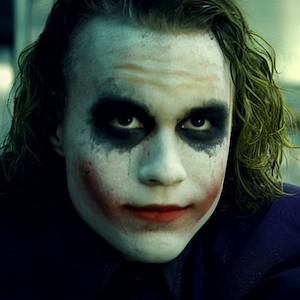} &
        \myim{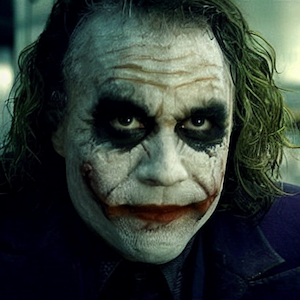} &
        \myim{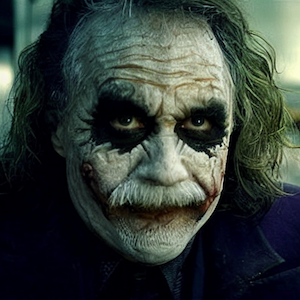} 
        & \myim{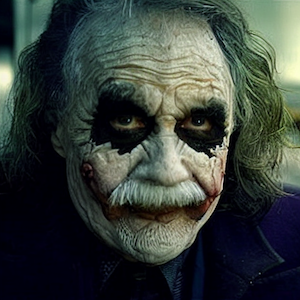} 
        
         \\

    \end{tabular}

   \caption{Supplementary qualitative results on images outside common face datasets}
   \label{fig:out_of_dist}
\end{figure}

To further evaluate the performance of FADING on images outside common face datasets, we conducted additional experiments on two out-of-distribution images, a publicly known figure and a movie character (Figure~\ref{fig:out_of_dist}).  Our method successfully achieved aging transformations on the image of the publicly known figure, demonstrating its generalization beyond dataset limitations. Furthermore, the model effectively handled the unique features and heavy makeup of the movie character, showcasing its adaptability to unconventional facial appearances. These experiments highlight the flexibility and robustness of~\method ~when applied to images beyond traditional datasets.

\end{document}